\documentclass[letterpaper]{article} 
\usepackage[draft]{aaai25}  
\usepackage{times}  
\usepackage{helvet}  
\usepackage{courier}  
\usepackage[hyphens]{url}  
\usepackage{graphicx} 
\usepackage{natbib}  
\usepackage{caption} 
\usepackage{algorithm}
\usepackage{algorithmic}
\usepackage{caption}

\usepackage[utf8]{inputenc} 
\usepackage[T1]{fontenc}    
\usepackage{url}            
\usepackage{booktabs}       
\usepackage{amsfonts}       
\usepackage{nicefrac}       
\usepackage{microtype}      
\usepackage{xcolor}         
\usepackage{makecell}
\usepackage{adjustbox}
\usepackage{pifont}
\usepackage{bbding}
\usepackage{amsmath}
\usepackage{multirow} 

\usepackage{newfloat}
\usepackage{listings}
\DeclareCaptionStyle{ruled}{labelfont=normalfont,labelsep=colon,strut=off} 
\floatstyle{ruled}
\newfloat{listing}{tb}{lst}{}
\floatname{listing}{Listing}
\title{Point Cloud Mamba: Point Cloud Learning via State Space Model}
\author{
    Tao Zhang\textsuperscript{\rm 1,2} \thanks{This work was performed when Tao Zhang was an Intern at Skywork AI. \textsuperscript{$\dagger$} Project Leader.}
    Haobo Yuan\textsuperscript{\rm 1}
    Lu Qi\textsuperscript{\rm 3}
    Jiangning Zhang\textsuperscript{\rm 4}
    Qianyu Zhou\textsuperscript{\rm 5} \\
    Shunping Ji\textsuperscript{\rm 1} 
    Shuicheng Yan\textsuperscript{\rm 2}
    Xiangtai Li\textsuperscript{\rm 2} \textsuperscript{$\dagger$} \\
}
\affiliations{
    \textsuperscript{\rm 1} Wuhan University
    \textsuperscript{\rm 2} Skywork AI 
    \textsuperscript{\rm 3} UC Merced \\
    \textsuperscript{\rm 4} Youtu Lab, Tencent
    \textsuperscript{\rm 5} Shanghai Jiao Tong University \\
zhang\_tao@whu.edu.cn, xiangtai94@gmail.com \\
Project page: \url{https://github.com/zhang-tao-whu/PCM}
}

\usepackage{bibentry}

\begin{document}

\maketitle

\begin{abstract}
Recently, state space models have exhibited strong global modeling capabilities and linear computational complexity in contrast to transformers.
%
This research focuses on applying such architecture to more efficiently and effectively model point cloud data globally with linear computational complexity.
In particular, for the first time, we demonstrate that Mamba-based point cloud methods can outperform previous methods based on transformer or multi-layer perceptrons (MLPs).
To enable Mamba to process 3-D point cloud data more effectively, we propose a novel Consistent Traverse Serialization method to convert point clouds into 1-D point sequences while ensuring that neighboring points in the sequence are also spatially adjacent. 
Consistent Traverse Serialization yields six variants by permuting the order of \textit{x}, \textit{y}, and \textit{z} coordinates, and the synergistic use of these variants aids Mamba in comprehensively observing point cloud data. 
Furthermore, to assist Mamba in handling point sequences with different orders more effectively, we introduce point prompts to inform Mamba of the sequence's arrangement rules. 
Finally, we propose positional encoding based on spatial coordinate mapping to inject positional information into point cloud sequences more effectively. 
Point Cloud Mamba surpasses the state-of-the-art (SOTA) point-based method PointNeXt and achieves new SOTA performance on the ScanObjectNN, ModelNet40, ShapeNetPart, and S3DIS datasets.
It is worth mentioning that when using a more powerful local feature extraction module, our PCM achieves 79.6 mIoU on S3DIS, significantly surpassing the previous SOTA models, DeLA and PTv3, by 5.5 mIoU and 4.9 mIoU, respectively.
%
\end{abstract}

\begin{figure}[t]
\centering
\includegraphics[width=0.46\textwidth]{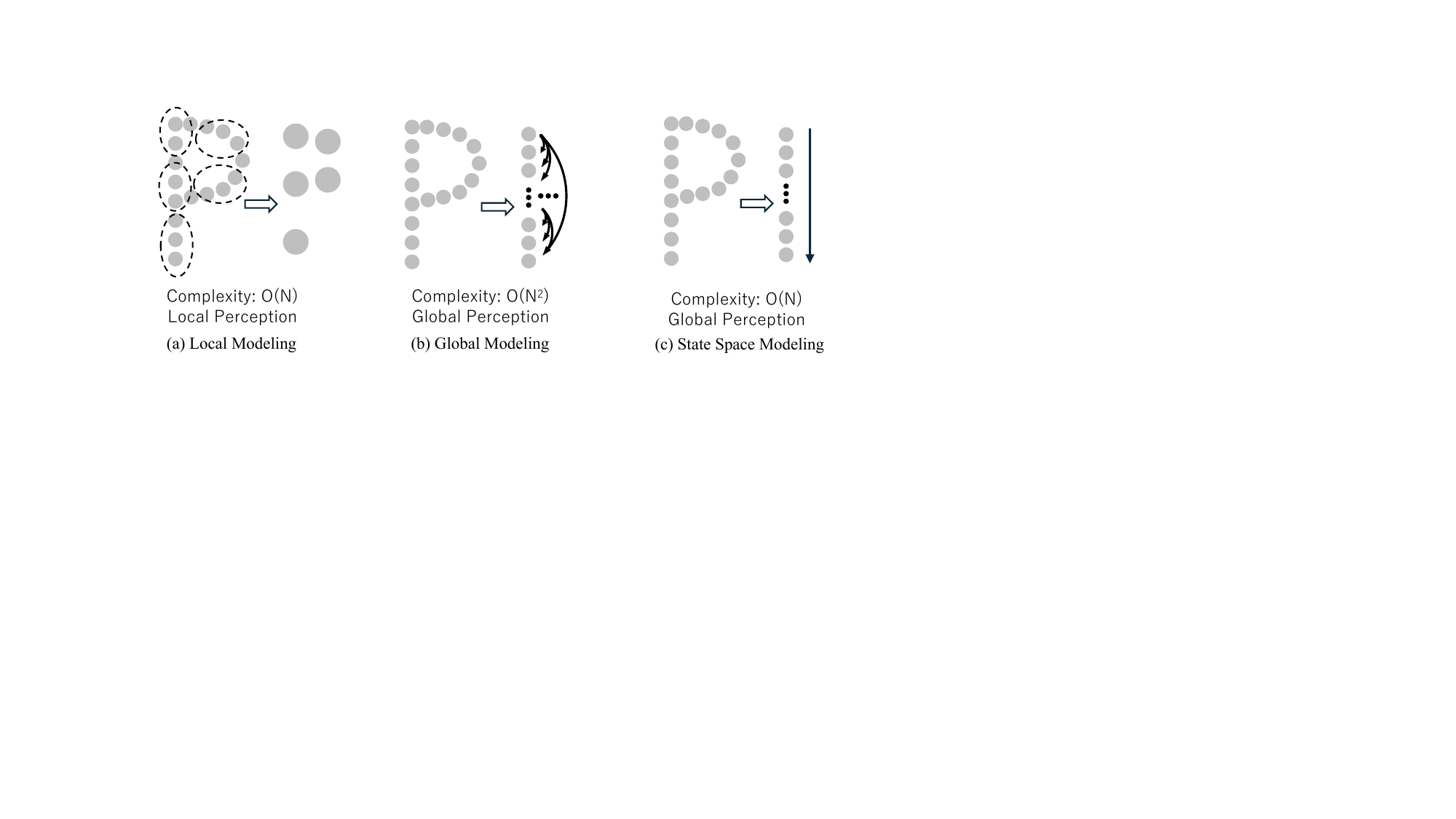}
\caption{\textbf{Several pipelines of point cloud modeling.} (a) denotes point-based methods with only local perception, including point-based methods, such as PointNet~\cite{qi2017pointnet}, PointNet++~\cite{qi2017pointnet++}, PointMLP~\cite{ma2022pointmlp}, and PointNeXt~\cite{qian2022pointnext}. (b) is the transformer-based method with global perception but quadratic computational cost, including Point Transformer~\cite{point_transformer} and Point-MAE~\cite{pang2022point-mae}. (c) represents Mamba-based methods, which offer advantages of global modeling and linear computational complexity.}
\label{fig:teaser}
\end{figure}
\section{Introduction}
\label{sec:intro}

Point cloud analysis~\cite{qi2017pointnet, qi2017pointnet++, li2018pointcnn, point_transformer, qian2022pointnext, wu2019pointconv, qian2022pointnext} has become a popular topic in 3D understanding and has drawn attention from the research community. 
Unlike 2D image processing, point clouds are composed of unordered and irregular point sets, making it difficult to apply the 2D image processing methods directly.
Thus, recent deep-learning-based approaches~\cite{qi2017pointnet, qi2017pointnet++, li2018pointcnn, wu2019pointconv, ma2022pointmlp, qian2022pointnext, point_transformer, guo2021pct} propose using various methods, such as voxel-based and point-based, for point cloud representation. 
The representative works, Point-Net~\cite{qi2017pointnet} and Point-Net++~\cite{qi2017pointnet++}, adopt the MLP-based design with deep hierarchical local priors, as shown in Figure~\ref{fig:teaser} (a). 
After that, many research works~\cite{ma2022pointmlp,boulch2020convpoint,  shi2020point, wang2019graph, point_transformer, guo2021pct, wu2022ptv2, wu2023ptv3} focus on advanced local geometric modeling via convolution, graph modeling, or attention.

Meanwhile, with the rapid progress of vision transformers, several works~\cite{point_transformer, guo2021pct, wu2022ptv2, wu2023ptv3} enhance the global modeling with transformer structure in point could, as shown in Figure~\ref{fig:teaser} (b). In addition, transformer architectures also work effectively in mask point pre-training, 3D segmentation, and in-context learning. However, the computation and memory costs are still huge. Recently, state space models~\cite{gu2023mamba,gu2022efficiently} (SSMs) have been proven to model long-range dependency in sequential data. In particular, Mamba~\cite{gu2023mamba} is proven effective as Transformer~\cite{vaswani2017attention} for several challenging NLP tasks. After that, recent works~\cite{zhu2024vision,liu2024vmamba,li2024mamba,behrouz2024GMN,U-Mamba,xing2024segmamba,yang2024vivim,ruan2024vm,he2024pan} explore SSMs in various vision tasks, including image representation learning, medical segmentation, and low-level vision tasks. One concurrent work, PointMamba, uses the Mamba layer to model the global context. 
However, there are still significant performance \textit{gaps} between PointMamba and previous point-based methods.

In this work, we ask an essential question: Can we design an efficient point cloud analysis architecture using Mamba and surpass the performance of point-based and transformer-based methods? In particular, we introduce the Point Cloud Mamba (PCM), a combining local and global modeling framework that outperforms the SOTA point-based and transformer-based methods.

PCM utilizes Mamba architecture to model the global features of point clouds while maintaining linear computational complexity. 
To process 3-D point cloud data effectively by Mamba layers, we propose a novel Consistent Traverse Serialization (CTS) method to serialize point clouds into a 1-D point sequence while ensuring that neighboring points in the sequence are also adjacent in space. 
Then, CTS can easily derive six variants by simply permuting the order of \textit{x}, \textit{y}, and \textit{z} coordinates. 
Additionally, when these six variants of CTS are combined, Mamba layers can more effectively model point cloud features.
This is because the different variants provide various perspectives of the point cloud, ensuring that spatially adjacent points are also adjacent in a serialized point sequence.
To help Mamba handle specific point sequences better, we introduce \textit{order} prompts to provide Mamba with the arrangement rules of the current point sequence. 
Finally, we propose simple spatial coordinate mapping as positional embedding for points, more suitable for irregular point cloud data than RoPE~\cite{su2024roformer} and learnable embedding.

Thanks to the above improvements, we successfully introduced Mamba into Point Cloud analysis and obtained Point Cloud Mamba (PCM). PCM outperforms the SOTA point-based method PointNeXt on three datasets: ScanObjectNet~\cite{uy2019scanobjectnn}, ModelNet40~\cite{wu2015ModelNet40}, and ShapeNetPart~\cite{yi2016shapenetpart}.
When enhancing the local feature extraction layers, PCM achieved 79.6 mIoU on the S3DIS dataset, significantly surpassing the previous SOTA PTv3~\cite{wu2023ptv3} by 4.9 mIoU.

In summary, we have the following contributions: 1) We introduce Mamba into point cloud analysis and construct a combined local and global modeling framework named Point Cloud Mamba. 
2) We propose consistent traverse serialization, order prompts, and positional encoding based on spatial coordinate mapping to assist Mamba in better handling point cloud data. 3) Point Cloud Mamba is the first Mamba-based method that works well in point cloud analysis. It outperforms the SOTA point-based method PointNeXt and transformer-based method PTV3 on ScanObjectNet, ModelNet40, ShapeNetPart, and S3DIS datasets.

\section{Related Work}
\label{sec:related_work}

\noindent
\textbf{3D Point Cloud Classification.} Recent works have used deep neural networks to process 3D point clouds. In particular, representative works, PointNet~\cite{qi2017pointnet} and PointNet++~\cite{qi2017pointnet++}, are the pioneering point-based approaches to handle the point clouds using MLPs directly. Meanwhile, several works explore graph-based modeling to utilize 3D geometric topology. Then, several works~\cite{wu2019pointconv, thomas2019kpconv,shen2018kcnet,li2018pointcnn,xu2021paconv,komarichev2019acnn,graham20183d,choy20194d,zhu2021cylindrical} explore the local geometric features via different kernel modeling. Moreover, several works~\cite{liu2019rscnn,xu2021GDANet,xiang2021curvenet,ran2022surface,chen2023pointgpt,jiang2022mae3d,liu2022masked, xie2020pointcontrast,zhang2021DepthContras,ma2022pointmlp,lang2019pointpillars,dai20183dmv,yan20222dpass} explore other point cloud architecture designs, including MLPs and transformers. Several works also explore different pre-training methods~\cite{pang2022point-mae,yu2021pointbert,sanghi2020info3d,sauder2019self,wang2019deep,hou2022point,jiang2023self,wu2023masked,zhu2023ponderv2,yang2023swin3d} or in-context abilities~\cite{fang2023explore,wang2023skeleton,wu2023towards}
inspired by the NLP field. Recently, state space models~\cite{gu2022efficiently,gu2023mamba} have achieved significant progress. Compared with transformers, they have advantages in efficient global modeling. A concurrent work~\cite{liang2024pointmamba} explores such architecture in point clouds. However, there are still significant performance gaps compared with previous point cloud methods. Our PCM shows that Mamba architecture can achieve comparable or even better results than transformer-based models.

\noindent
\textbf{3D Visual Transformers.} With the rise of the transformer in 2D version~\cite{dosovitskiy2020imageVIT,li2023transformer,detr}, several works~\cite{lahoud20223d,guo2021pct,Schult23ICRA,sun2022superpoint,lai2022stratified,liu2023flatformer,wang2023octformer} also explore transformer architectures in the point cloud. Earlier works~\cite{guo2021pct,point_transformer} have focused on the point cloud process. PCT~\cite{guo2021pct} performs global attention directly to each point, following the ViT~\cite{dosovitskiy2020imageVIT}. However, it has memory consumption and computational complexity issues. Point Transformer~\cite{point_transformer} solves this issue by introducing local attention. Then, the updated versions~\cite{wu2022ptv2,wu2023ptv3} explore the different architectures to improve performance and efficiency. 
Inspired by these studies, our works combine local point processing and a new traverse serialization strategy, which leads to better results than direct SSM traverse.

Transformer-based methods~\cite{point_transformer,guo2021pct,wu2022ptv2,wu2023ptv3,lai2022stratified,robert2023efficient,park2023self,wang2023octformer} use Transformers to model point cloud sequences and have extensively explored a lot of aspects, such as point cloud serialization~\cite{wang2023octformer,point_transformer,guo2021pct,wu2022ptv2,wu2023ptv3} and point positional embedding~\cite{wu2022ptv2,park2023self}. Although it's possible to simply implement a Mamba-based network for point cloud analysis by replacing Transformer layers with Mamba layers, due to the differences between Transformers and Mamba, substantial exploration is still needed to find the most suitable network architecture, serialization method and positional embedding strategy for Mamba-based models. This is precisely the objective of this paper.\looseness=-1

\noindent
\textbf{State Space Models.} Inspired by continuous state space models in control systems, recently, state space models~\cite{gu2023mamba,gu2022efficiently} have been proven to model long-range dependency. In particular, S4~\cite{gu2022efficiently} proposes to normalize the parameter into the diagonal structure, which results in less computation cost and memory usage. After that, Mamba~\cite{gu2023mamba} presents a selection mechanism that leads to better results than transformers. Recently, several works have explored such architecture in different tasks, including image classification~\cite{zhu2024vision,liu2024vmamba,li2024mamba}, graph modeling~\cite{behrouz2024GMN}, medical segmentation~\cite{U-Mamba,xing2024segmamba,yang2024vivim,ruan2024vm}, and low-level version tasks~\cite{he2024pan}. As a concurrent work, we further prove the potential of SSMs in the 3D point clouds, where we can achieve even better results than previous architectures.
\section{Method}
\label{sec:method}
\begin{figure}[t]
\centering
\includegraphics[width=0.46\textwidth]{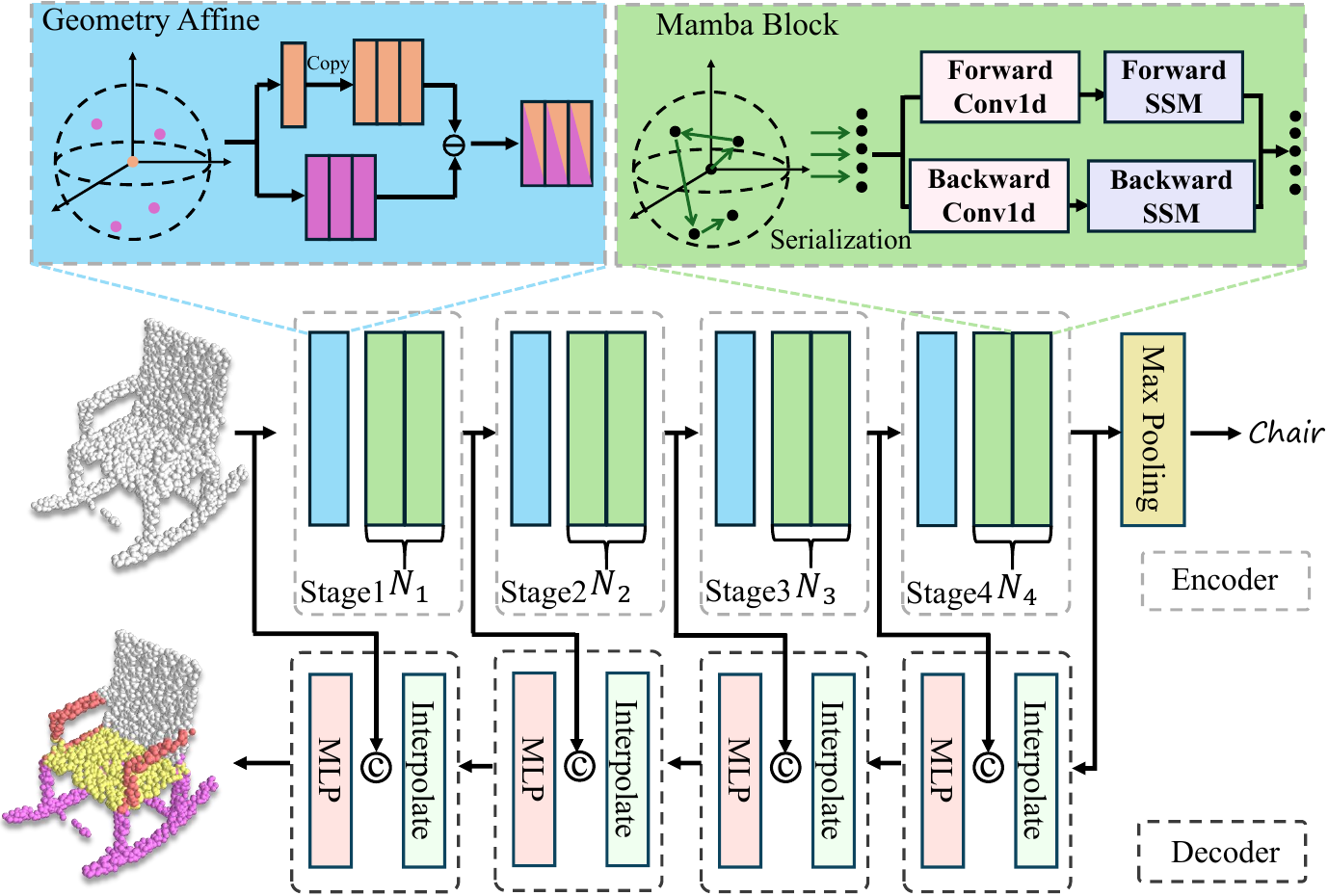}
\caption{\textbf{The architecture of our proposed Point Cloud Mamba.} PCM encoder consists of four stages, each comprising a geometric affine module and several mamba layers. Point downsampling is performed between stages. The decoder only consists of point interpolation, feature concatenation, and MLP.}
\label{fig:method}
\end{figure}

%
The SSM-based architecture, Mamba~\cite{gu2023mamba}, is attractive for point cloud representation learning due to its global modeling capability and linear computational complexity. However, Mamba is designed for the causal modeling of 1-D sequences, making it difficult to be directly applied to the modeling of non-causal 3-D point cloud data, posing many challenges to be addressed. This section explores how to effectively integrate Mamba into architectures based on local modeling to capture global features.


\subsection{Preliminaries}
\label{sec:preliminary}

\noindent
\textbf{PointMLP formulation.}
PointMLP~\cite{ma2022pointmlp} models the representation of point clouds using simple MLP and expands the receptive field of each point through point cloud downsampling and local feature aggregation. The process can be described by Equ.~\ref{eq1}, \ref{eq2}, and \ref{eq2_2}:
\begin{equation}
      f_{i}^{l+1} = \Phi_{2}(\mathcal{A}(\Phi_{1}(GAM(\{f_{i,j}^{l}\})), | j = 1,...,K)),
\label{eq1}
\end{equation}
\begin{equation}
      GAM(\{f_{i,j}^{l}\}) = \alpha \odot \frac{\{f_{i,j}^{l}\}-f_{i}^{l}}{\sigma + \delta},
\label{eq2}
\end{equation} 
\begin{equation}
      \sigma = \sqrt{\frac{1}{k \times n \times d} \sum_{i=1}^{n} \sum_{j=1}^{k} (f_{i,j}^{l} - f_{i}^{l})^{2}},
\label{eq2_2}
\end{equation}
where $\Phi$ represents a network composed of a series of residual MLPs, $\{f_{i,j} | j=1,...,k\}$ represents the K neighboring points of $f_{i}$, $\mathcal{A}$ denotes the max-pooling operation. GAM refers to the Geometric Affine Module proposed by PointMLP to enhance local features.


\noindent
\textbf{Mamba formulation.}
The state-space equation can describe a multi-input, multi-output continuous system where the current inputs and states jointly determine the change in the state space of this system:
\begin{equation}
      h^{'}(t) = \mathbf{A}h(t) + \mathbf{B}x(t),\quad y(t) = \mathbf{C}h(t),
\label{eq3}
\end{equation}
where \textit{x(t)}, \textit{h(t)}, and \textit{y(t)} are the inputs, states, and outputs of the current system, respectively. $\mathbf{A}$, $\mathbf{B}$, and $\mathbf{C}$ are all continuous parameters of the system.

The continuous state-space equation mentioned above can be transformed into a discrete formulation using a timescale parameter $\Delta$ based on the zero-order hold rule:
\begin{equation}
      \overline{\mathbf{A}} = exp(\Delta \mathbf{A}), \quad 
      \overline{\mathbf{B}} = (\Delta \mathbf{A})^{-1}(exp(\Delta \mathbf{A}) - I) \cdot \Delta \mathbf{B},
\label{eq4}
\end{equation}
\begin{equation}
      h_{t} = \overline{\mathbf{A}}h_{t-1} + \overline{\mathbf{B}}x_{t},\quad y_{t} = \mathbf{C}h_{t},
\label{eq5}
\end{equation}
where $x_t$, $h_t$, and $y_t$ are the system's discrete inputs, states, and outputs. $\overline{\mathbf{A}}$, $\overline{\mathbf{B}}$ are all discrete parameters of the system.

Inspired by Equ.~\ref{eq4} and \ref{eq5}, Mamba~\cite{gu2023mamba}, a new SSM-based model that introduces time-varying system parameters, has been introduced. Specifically, $\Delta$, $\overline{\mathbf{A}}$, and $\overline{\mathbf{B}}$ are all functions of $x_t$. Mamba demonstrates long sequence modeling capabilities comparable to Transformer and achieves linear computational complexity during inference following Equ.~\ref{eq5}. However, Equ.~\ref{eq5} is difficult to compute in parallel. It can be expanded and implemented using global convolution to enhance the efficiency of training Mamba on GPUs:
\begin{equation}
      \overline{\mathbf{K}} = (\mathbf{C}\overline{\mathbf{B}}, \mathbf{C}\overline{\mathbf{A}}\overline{\mathbf{B}}, ... , \mathbf{C}\overline{\mathbf{A}}^{M-1}\overline{\mathbf{B}}), \quad 
      y = x \ast \overline{\mathbf{K}},
\label{eq6}
\end{equation}
where $M$ is the length of the input sequence $x$, and $\overline{\mathbf{K}}$ is the kernel of the global convolution.

\subsection{A Naive Mamba-based Point Cloud Network}
\label{sec:mamba_point_net}

\noindent
\textbf{Architecture.} Using Mamba~\cite{gu2023mamba} to replace the MLP operator in PointMLP~\cite{ma2022pointmlp} to achieve global modeling capabilities is a natural idea. However, this still requires addressing two challenges: 1) Point clouds are 3-D data, so how can we transform them into 1-D sequences? 2) Mamba is designed for causal modeling, so how can Mamba handle non-causal point cloud data? To address these challenges, we first adopt the z-order~\cite{morton1966computer} serialization method~\cite{wang2023octformer} to flatten 3-D point cloud data into 1-D sequences, allowing point cloud data to be processed by Mamba. Secondly, inspired by \cite{zhu2024vision} and \cite{liu2024vmamba}, we use bidirectional Mamba to allow each point to obtain features from any other point. At this point, we have implemented a naive Mamba-based network for the point cloud:
\looseness=-1
\begin{equation}
      f_{i}^{l+1} = Mamba(x^{l}, reverse(x^{l})),
\label{eq7}
\end{equation}
\begin{equation}
      x^{l} = \mathcal{S}_{z}(\mathcal{A}(GAM(\{f_{i,j}^{l} | j=1,...,k\}))), 
\label{eq7_2}
\end{equation}
where $\mathcal{S}_{z}$ refers to serialization according to the z-order.

\noindent
\textbf{Shortcomings.} The naive Mamba-based network still has some shortcomings, only achieving 84.1 OA on ScanObjectNN, significantly underperforming modern point-based methods like PointNeXt. There are some reasons: 1) Using a single serialization method to convert 3-D point cloud data into 1-D point sequences significantly loses the spatial relational information between points. Specifically, even if a point is adjacent to many others in space, it can only be adjacent to adjacent points in the point sequence. Therefore, employing multiple serialization methods in combination to alleviate this loss of spatial relational information is necessary. 2) The architecture design of the naive Mamba-based network follows the point-based method PointMLP, which may not be optimal or even reasonable for Mamba-based architecture. Exploring what benefits Mamba in modeling point clouds is important and necessary.
%
%

\subsection{Point Cloud Mamba}
\label{sec:pcm}
In the last section, we obtained a naive mamba-based point cloud network; however, there is still plenty of room for optimization. Next, we will elaborate on improving this naive architecture to Point Cloud Mamba (PCM) and achieving performance beyond PointNeXt~\cite{qian2022pointnext} and PTv3~\cite{wu2023ptv3}. 

Firstly, PCM provides various point cloud sequences through the consistent traverse serialization strategy and its variants, thereby preserving the relational information between points to the fullest extent by traversing multiple point sequences. Secondly, PCM introduces order prompts to assist Mamba layers in better handling point sequences generated by different serialization methods and strengthens the spatial position information of points through positional encoding. Finally, we have designed a more reasonable overall architecture.

\begin{figure}[t]
\centering
\includegraphics[width=0.45\textwidth]{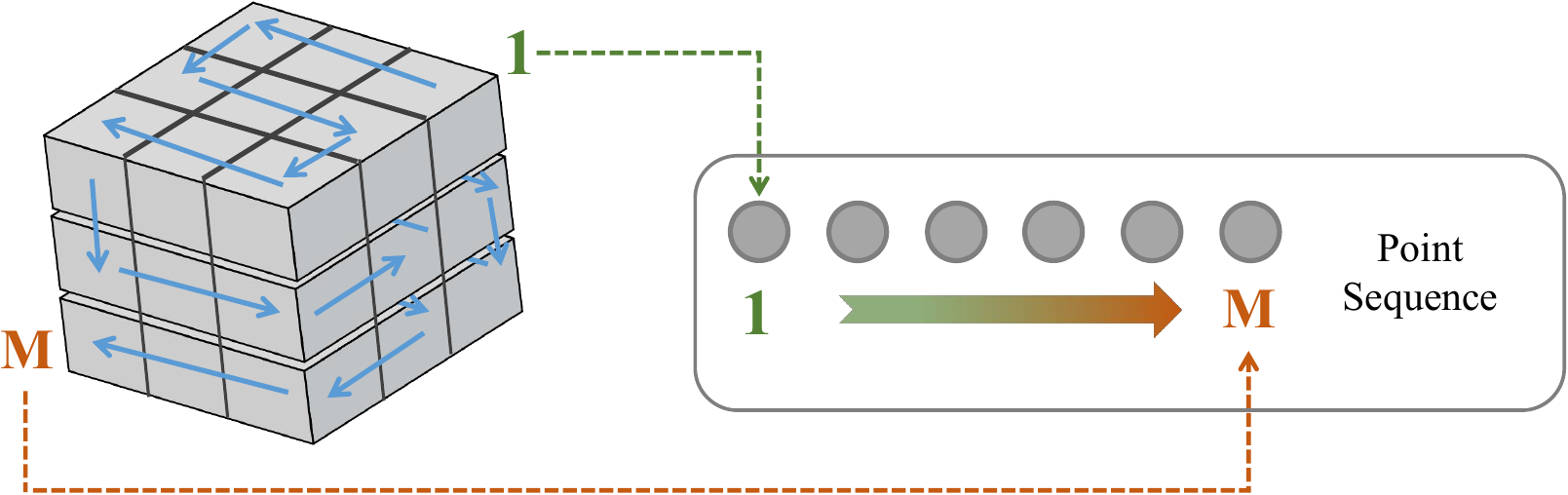}
\caption{\textbf{The consistent traverse serialization strategy.} The 3-D point cloud data is voxelized and then serialized into a 1-D point sequence according to a predefined order. M represents the total number of points in the point cloud. With the permutation of x, y, and z coordinates, consistent traverse serialization has six variants.}
\label{fig:cts}
\end{figure}

\noindent
\textbf{Serialization strategy.} How to convert 3-D point cloud data into a 1-D sequence that Mamba can handle is crucial. 
%
We find that Mamba can better process point cloud sequences arranged according to specific rules than disordered point cloud sequences. 
For example, randomly flattening point cloud data into a 1-D point sequence and feeding it into Mamba for modeling will significantly underperform compared to using an ordered point sequence (84.1 vs. 86.7).
%
Based on this, we propose the Consistent Traverse Serialization (CTS) strategy, which ensures that adjacent points in the sequence are also adjacent in spatial position. 
Figure~\ref{fig:cts} illustrates how CTS converts 3-D point clouds into a 1-D sequence. 

Firstly, the point cloud is grid sampled to transform continuous spatial coordinates into discrete grid coordinates:
\begin{equation}
      \{c_{1}^{g}, c_{2}^{g}, c_{3}^{g}\} = 
      int(\{c_{1}^{s} , c_{2}^{s}, c_{3}^{s}\} \times N), 
\label{eq8}
\end{equation}
where $\{c_{1}^{s}, c_{2}^{s}, c_{3}^{s}\}$ and $\{c_{1}^{g}, c_{2}^{g}, c_{3}^{g}\}$ are the spatial and grid coordinates of the points, and $N$ is the grid number. 
We design an encoding function that, given the coordinates of two dimensions, maps the coordinates to a code. Sorting the sequence according to the code ensures that adjacent points are contiguous in space.
\begin{small}
\begin{equation}
       Code\_func(n_{1}, n_{2}) = \left\{
    \begin{aligned}
         n_{2} \times N + n_{1},  & n_{2} \% 2 = 0 \\
         (n_{2} + 1) \times N - n_{1}, & n_{2} \% 2 \neq 0 \\
    \end{aligned}
    \right.
\label{eq9}
\end{equation}
\end{small}
Then, we can compute a code for each point based on its grid coordinates:
\begin{equation}
      \textbf{code} = Code\_func(Code\_func(c_{1}^{g}, c_{2}^{g}), c_{3}^{g}).
\label{eq10}
\end{equation}

Sorting the point cloud according to the $\textbf{code}$ allows it to be flattened into a 1-D sequence. 
This simple serialization strategy performs comparably to carefully designed z-order~\cite{morton1966computer} and Hilbert-order~\cite{hilbert1935stetige} serialization strategies~\cite{wang2023octformer, wu2023ptv3}. 
Additionally, by exchanging the order of the x, y, and z axes, six different serialization methods can be derived, which we call "xyz", "xzy", "yxz", "yzx", "zxy", and "zyx". 
These different serialization methods can be viewed as various point cloud observations from different spatial perspectives.

Additionally, combining multiple serialization strategies can effectively assist Mamba in better modeling point cloud features, as shown in Table~\ref{tab:order_prompts}. 
Specifically, we adopt different serialization strategies for the inputs of different Mamba layers, allowing Mamba to perceive the point cloud more comprehensively, thus significantly surpassing a single serialization strategy alone.

\begin{figure}[t]
\centering
\includegraphics[width=0.46\textwidth]{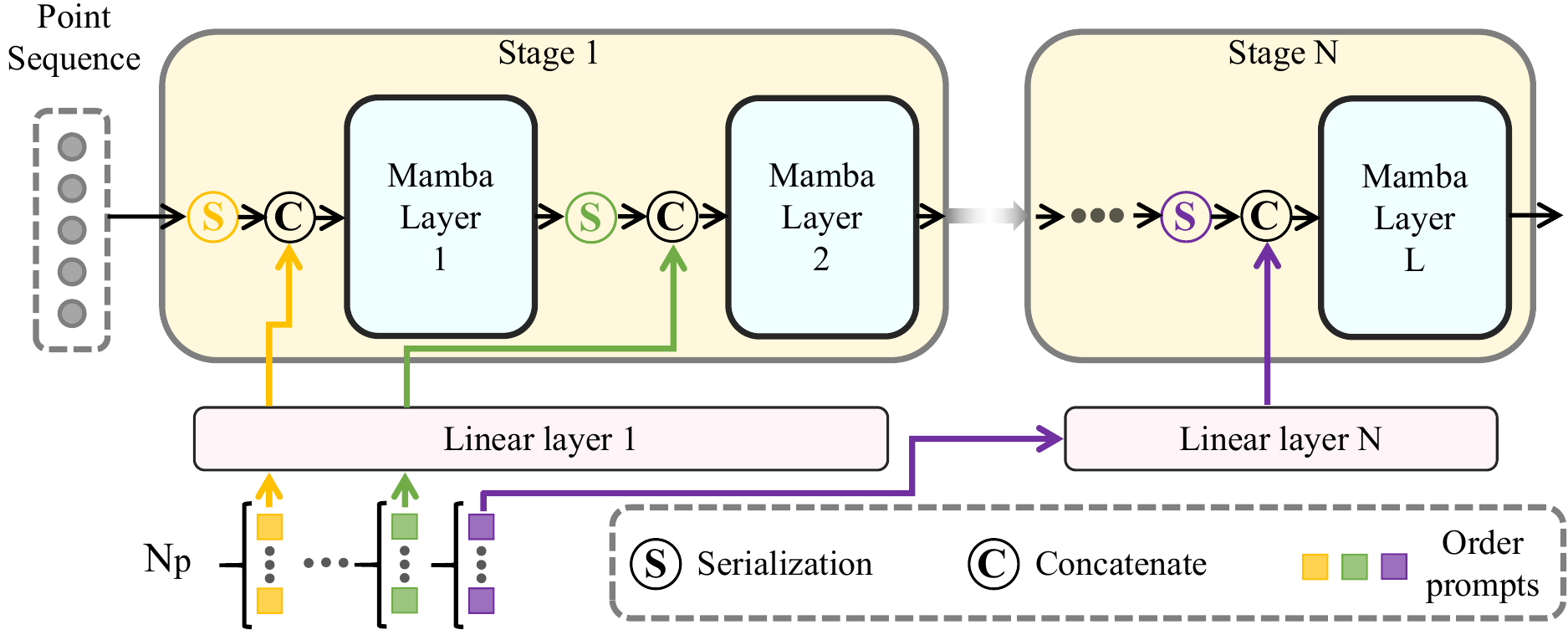}
\caption{\textbf{The order prompts.} Different colors represent different serialization orders. $N_p$ order prompts are mapped to the same channel size as the features and then concatenated to the beginning and end of the input point sequence.}
\label{fig:order_prompt}
\end{figure}

\noindent
\textbf{Order prompt.} When multiple serialization strategies are combined, assigning an identifier to each serialization is necessary. The identifier can help the Mamba layers recognize the arrangement of point cloud sequences and better capture point cloud features. We propose a simple but efficient order prompt mechanism to achieve this goal, which resembles the system messages in large language models.

As shown in Figure~\ref{fig:order_prompt}, we assigned $N_p$ learnable embeddings as order prompts for each serialization order. Before being processed by the Mamba layer, the point cloud sequence has the corresponding order prompts added to both the beginning and end of the sequence. Considering that the input feature dimensions of the Mamba layers at different stages may vary, we also allocate a shared Linear layer for all Mamba layers within each stage to map the order prompts to the required channel size.

\noindent
\textbf{Positional embedding.} 
In sequence modeling, positional encoding is crucial and widely applied in text and image patch sequence modeling. 
For example, RoPE~\cite{su2024roformer} and learnable positional embedding are widely used in text sequence modeling and image patch sequence modeling, respectively. 
However, we find that these positional encoding methods are unsuitable for point clouds due to their sparsity and irregular shapes. The spatial distances between any two adjacent points in a point cloud sequence vary significantly, making it difficult to use the sequence positional information to represent the spatial gaps between adjacent points accurately. 
We employ a shared positional mapping function to address this challenge. 
It maps the point coordinates to positional embeddings with the same channel size as the features.
\looseness=-1
\begin{equation}
      Emd_{pos} = Linear(\{c_{1}^{s}, c_{2}^{s}, c_{3}^{s}\}). 
\label{eq11}
\end{equation}
$Emd_{pos}$ refers to the positional embedding projected from the spatial coordinates of the point. This simple positional mapping function accurately encodes the spatial information of points into features, and the positional embeddings corresponding to adjacent points in the sequence are similarly similar. Since the channel size of features varies across different stages of the network, we need to learn a private positional embedding function for each stage, and different Mamba layers within the same stage share the same positional encoding function.

\begin{table}[t]
\centering
\resizebox{0.47\textwidth}{!}{
\begin{tabular}{ l | c | c }
\toprule
  Architecture & PCM-Tiny & PCM \\
\midrule
Mamba layers &\{1, 1, 2, 2\} & \{1, 2, 2, 4\} \\
Serialization & \{[xyz]-[xzy]-[yxz, yzx]-[zxy-zyx]\} & \{[xyz]-[xzy, yxz]-[yzx, zxy]-[zyx, H, z, z-trans]\} \\
Channels & \{192, 192, 384, 384\} & \{384, 384, 768, 768\} \\
Order Prompts & 6 & 6\\
\bottomrule
\end{tabular}
}
\caption{\textbf{Architecture settings.} The parameters of the four stages are enclosed in curly braces \{~\}, while the parameters corresponding to each Mamba layer within a stage are enclosed in square brackets [~].}
\label{tab:architecture_settings}
\end{table}

\noindent
\textbf{Architecture settings.} As shown in Figure~\ref{fig:method}, our network architecture consists of four stages, each incorporating a geometric affine module and several Mamba layers. 
We employ a simple decoder without Mamba layers for point cloud segmentation. The decoder solely conducts point cloud interpolation, concatenation with multi-stage encoder features, and channel transformation through MLP.
The number of Mamba layers, serialization strategies, channel sizes, and order prompt counts in each stage of PCM-Tiny are illustrated in Table~\ref{tab:architecture_settings}. 
Furthermore, we obtain PCM by increasing the number of Mamba layers and channel sizes and adopting new serialization methods accordingly.

\section{Experiments}
\label{sec:exp}




\begin{table}[t]
\centering
\resizebox{0.47\textwidth}{!}{
\begin{tabular}{ l | l l | l l | c }
\toprule
 & \multicolumn{2}{c|}{\textbf{ScanObjectNN (PB\_T50\_RS)}} & \multicolumn{2}{c|}{\textbf{ModelNet40}}&  Params. \\
\textbf{Method}  & OA (\%) & mAcc (\%) & OA (\%) & mAcc (\%) & M \\
\midrule
PointNet~\cite{qi2017pointnet}&68.2  &63.4  & 89.2 & 86.2 & 3.5 \\ 
PointCNN~\cite{li2018pointcnn}&78.5  &75.1 & 92.2& 88.1 & 0.6 \\
KPConv~\cite{thomas2019kpconv}  & -& -& 92.9 & - & 14.3\\ 
ASSANet-L ~\cite{qian2021assanet}  & - & -& 92.9 & - & 118.4\\ 
CurveNet~\cite{xiang2021curvenet}   & - & -& 93.8 & - & 2.0\\ 
PointMLP~\cite{ma2022pointmlp}  & 85.4$\pm$1.3 & 83.9$\pm$1.5 & \textbf{94.1} & \textbf{91.3} & 13.2 \\
PointNet++~\cite{qi2017pointnet++} & 77.9 &75.4 & 91.9& - & 1.5  \\
PointNeXt~\cite{qian2022pointnext} & 87.7$\pm$0.4 & 85.8$\pm$0.6 & 93.2$\pm$0.1 & 90.8$\pm$0.2 & 1.4 \\
\midrule
\multicolumn{6}{c}{Transformer-based} \\
\midrule
PCT~\cite{guo2021pct}  & - & -& 93.2& -  & 2.9 \\
Point-BERT~\cite{yu2021pointbert} & 83.1 & - & 93.2 & - & 22.1 \\
Point-MAE~\cite{pang2022point-mae} & 85.2 & - & 93.8 & - & 22.1 \\
PTv3~\cite{wu2023ptv3} & 86.4 & 83.9 & - & - & - \\
\midrule
\multicolumn{6}{c}{Mamba-based} \\ 
\midrule
PointMamba~\cite{liang2024pointmamba} & 84.9 & - & - & - & 12.3 \\
PCM-Tiny (ours) & 86.9$\pm$0.4 & 85.0$\pm$0.3 & 93.1$\pm$0.1 & 90.6$\pm$0.3 & 6.9 \\
PCM (ours) & \textbf{88.1}$\pm$0.3 & \textbf{86.6}$\pm$0.2 & 93.4$\pm$0.2 & 90.7$\pm$0.6 & 34.2 \\
\bottomrule
\end{tabular}
}
\caption{\textbf{3D object classification in ScanObjectNN and ModelNet40.} Averaged results in three random runs using $1024$ points as input without voting are reported.}
\label{tab:classificaition}
\end{table}

We conduct experiments on four datasets: ScanObjectNN~\cite{uy2019scanobjectnn} and ModelNet40~\cite{wu2015ModelNet40} classification datasets, ShapeNetPart~\cite{chang2015shapenet} part segmentation dataset, and S3DIS~\cite{armeni20163d} semantic segmentation dataset. For the detailed experiment settings, please refer to the supplementary materials.

\begin{table}[t]
\centering
\resizebox{0.47\textwidth}{!}{
\begin{tabular}{l|ll|c}
\toprule
\textbf{Method}  & ins.\ mIoU  & cls. \ mIoU & Params. \\ %
\midrule
PointNet~\cite{qi2017pointnet} & 83.7 & 80.4 & 3.6 \\
CurveNet~\cite{xiang2021curvenet} & 86.8 & - & -  \\
ASSANet-L~\cite{qian2021assanet} & 86.1 & - & -  \\
Point Transformer~\cite{point_transformer} & 86.6 & 83.7 & 7.8 \\ 
PointMLP~\cite{ma2022pointmlp} & 86.1 & 84.6 & - \\ 
PointNet++~\cite{qi2017pointnet++} & 85.1 & 81.9 & 1.0 \\
PTv1~\cite{point_transformer} & 86.6 & 83.7 & -  \\
PointNeXt-S (C=160)~\cite{qian2022pointnext} & 86.5$\pm$0.1 & - & 22.5 \\
DeLA~\cite{chen2023decoupled} & 87.0 & 85.8 & 7.5 \\
SpoTr~\cite{park2023self} & 87.2 & 85.4 & - \\
\midrule
PointMamba~\cite{liang2024pointmamba} & 86.0 & 84.4  & 17.4 \\
PCM-Tiny (ours) & 86.9 & 85.0 & 8.8 \\
PCM (ours) & 87.0$\pm$0.2 & 85.3$\pm$0.1 & 40.6\\
\bottomrule
\end{tabular}
}
\caption{\textbf{Part segmentation in ShapeNetPart.}}\vspace{-2mm}
\label{tab:shapenetpart}
\end{table}

\begin{table}[t]
\centering
\resizebox{0.47\textwidth}{!}{
\begin{tabular}{l|ccc}
\toprule
\textbf{Method}  & OA & mAcc & mIOU\\ %
\midrule
PointNet~\cite{qi2017pointnet} & - & 49.0 & 41.1  \\
PointCNN~\cite{li2018pointcnn} & 85.9 & 63.9 & 57.3 \\
PointNeXt~\cite{qian2022pointnext} & 91.0 & 77.2 & 71.1 \\
Strat. Trans.~\cite{lai2022stratified} & 91.5 & 78.1 & 72.0 \\
SPT~\cite{robert2023efficient} & - & - & 68.9 \\
SpoTr~\cite{park2023self} & 90.7 & 76.4 & 70.8 \\

PTv1~\cite{point_transformer} & 90.8 & 76.5 & 70.4 \\
PTv2~\cite{wu2022ptv2} & 91.6 & 78.0 & 72.7 \\
DeLA~\cite{chen2023decoupled} & 92.2 & 80.1 & 74.1 \\
DeLA+X-3D~\cite{sun2024x} & 92.2 & 80.1 & 74.3 \\
KPConvX-L~\cite{thomas2024kpconvx} & 91.7 & 78.7 & 73.5 \\
PTv3~\cite{wu2023ptv3} & - & - & 74.7 \\
PTv3\dag~\cite{wu2023ptv3} & 91.4 & 78.4 & 72.3 \\
\midrule
PCM-Tiny~(ours) & 92.9 & 81.6 & 74.1 \\ 
PCM-Tiny\dag~(ours) & \textbf{95.1} & \textbf{82.8} & \textbf{79.6}\\
\bottomrule
\end{tabular}
}
\caption{\textbf{3D semantic segmentation in S3DIS.} \dag~indicates using DeLA~\cite{chen2023decoupled} blocks as the additional local feature extractor.}\vspace{-4mm}
\label{tab:s3dis}
\end{table}

\subsection{Main Results}

\noindent
\textbf{3D object classification in ScanObjectNN dataset.}
ScanObjectNet~\cite{uy2019scanobjectnn} is a challenging point cloud classification dataset containing 15,000 real scanned objects categorized into 13 classes. It is known for its noise and occlusion challenges. Following PointMLP~\cite{ma2022pointmlp} and PointNeXt~\cite{qian2022pointnext}, we conducted experiments on PB\_T50\_RS, the most challenging and commonly used ScanObjectNN variant. As shown in Table~\ref{tab:classificaition}, PCM achieved an OA of 88.1 and a mAcc of 86.6 on ScanObjectNN, surpassing the SOTA method PointNeXt and PTv3 by 0.4 and 1.7 in OA, respectively. Compared to PointMLP, Mamba layers demonstrated significantly stronger modeling capabilities than MLP, with PointNeXt surpassing PointMLP by 2.7 in OA and 2.7 in mAcc. By reducing the number of Mamba layers and channel size, PCM-Tiny achieved an OA of 86.9 and a mAcc of 85.0 with only 20\% of the parameters of PCM. It is worth noting that PCM-Tiny, with only 52\% of the parameters of PointMLP (6.9 M vs. 13.2 M), still outperformed PointMLP by 1.5 in OA and 1.1 in mAcc.

The superior performance of PCM compared to PointMLP demonstrates the importance of Mamba's global modeling capability for point cloud analysis.

\noindent
\textbf{3D object classification in ModelNet40 dataset.} ModelNet40~\cite{wu2015ModelNet40} is a widely used synthetic 3D object classification dataset consisting of 40 categories, each with 100 unique CAD models. As shown in Table~\ref{tab:classificaition}, PCM achieved an OA of 93.4 and a mAcc of 90.7, reaching a performance comparable to PointNeXt~\cite{qian2022pointnext}. PCM-Tiny achieved an OA of 93.1 and a mAcc of 90.6 with approximately 20\% of the parameters of PCM. However, due to the smaller scale and less challenging nature of ModelNet40, performance on this dataset is difficult to differentiate between different methods' modeling capabilities significantly, with most methods' OA concentrated between 93 and 94. We have reproduced the experiments of PointMLP~\cite{ma2022pointmlp} and obtained an OA of 93.6, indicating that the high accuracy of 94.1 requires multiple repetitions and selection of the best.

\noindent
\textbf{3D object part segmentation in ShapeNetPart dataset.} ShapeNetPart~\cite{chang2015shapenet} is a widely used dataset for 3D object part segmentation. It comprises 16,880 models from 16 different shape categories and 50 part labels. Experimental results on the ShapeNetPart dataset are shown in Table~\ref{tab:shapenetpart}. PCM achieves 87.0 Ins. mIoU and 85.3 Cls. mIoU without using extra test augmentation strategies such as voting, surpassing PointNeXt~\cite{qian2022pointnext} by 0.5 Ins. mIoU. Point-Tiny achieves 86.9 Ins. mIoU and 85.0 Cls. mIoU, surpassing PointNeXt by 0.4 Ins. mIoU. PCM outperforms PointMLP~\cite{ma2022pointmlp} by 1.0 Ins. mIoU and 1.0 Cls. mIoU, demonstrating the significant potential of Mamba for 3D point cloud modeling.

\noindent
\textbf{3D semantic segmentation in S3DIS dataset.} S3DIS~\cite{armeni20163d} is a large-scale indoor point cloud benchmark containing 6 large indoor areas, 271 rooms, and 13 semantic categories. PCM achieved 74.1 mIoU and 92.9 OA, surpassing PointNext~\cite{qian2022pointnext} by 3.0 mIoU and 1.9 OA and exceeding PTv2~\cite{wu2022ptv2} by 1.4 mIoU and 1.3 OA. Moreover, PCM attained performance comparable to the current SOTA point-based method DeLA~\cite{chen2023decoupled} and transformer-based method PTv3~\cite{wu2023ptv3}.

For better extraction of local point features, we cascade 4 DeLA blocks before PCM as an additional local feature extractor; please refer to the supplementary for details. When combining with the more powerful local feature extractor~\cite{chen2023decoupled}, PCM-Tiny achieved 95.1 OA, 82.8 mAcc, and 79.6 mIoU, significantly surpassing the previous SOTA models DeLA~\cite{chen2023decoupled} and PTv3~\cite{wu2023ptv3} by 5.5 mIoU and 4.9 mIoU, respectively.

We also evaluate the performance of PTv3 with the same additional DeLA local feature extractor. However, the additional local feature extractor does not bring performance improvements to PTv3. This might be due to PTv3 performing attention within a window size of 1024, a limitation imposed by the quadratic computational complexity of transformers, which results in its limited global modeling capability.

\begin{table}[t]
\centering
\resizebox{0.47\textwidth}{!}{
\begin{tabular}{l|cc}
        \toprule
        \textbf{Strategy}  & OA (\%) & mAcc (\%) \\ %
        \midrule
        \{"z"\}$\times$ 9  & 86.78 & 84.67 \\
        \{"hilbert"\} $\times$ 9 & 86.78 & 84.68 \\
        \{"xyz"\} $\times$ 9 & 86.71 & 85.00 \\
        \midrule
        \{"xyz", "yzx", "zxy"\} $\times$ 3 & 86.88 & 85.11 \\
        \{"xyz", "xzy", "yxz", "yzx", "zxy", "zyx", "xyz", "yzx", "zxy"\} & 87.10 & 85.51 \\
        \{"xyz", "xzy", "yxz", "yzx", "zxy", "zyx", "hilbert", "z", "z-trans"\} & 87.20 & 85.54 \\
        \bottomrule
        \end{tabular}
}
\caption{\textbf{Ablation studies on serialization strategies.} Each mamba layer is assigned a serialization order and listed inside \{\}. "xyz", "xzy", "yxz", "yzx", "zxy",  and "zyx" represent different variants of our proposed Consistent Traverse Serialization strategy.}\label{tab:serialization}
\end{table}

\begin{table}[t]
\centering
\resizebox{0.45\textwidth}{!}{
\begin{tabular}{l|cc|c}
        \toprule
        \textbf{Channels
        }  & OA (\%) & mAcc (\%) & Params. (M) \\%
        \midrule
        \{96-96-96-96\} & 84.84 & 82.11 & 1.2 \\
        \{192-192-192-192\} & 85.91	& 84.35 & 3.7 \\
        \{384-384-384-384\} & 86.40 & 84.48 & 12.7 \\
        \{768-768-768-768\} & 87.52 & 85.87 & 47.2\\
        \midrule
        \{96-192-384-768\} & 86.16 & 83.67 & 22.6\\
        \{384-384-768-768\} & 87.40 & 85.52 & 34.2 \\
        \bottomrule
        \end{tabular}
}
\caption{\small{\textbf{Ablation studies on channel size.} The four-stage feature channel sizes are listed inside \{\} and connected with -.}}\label{tab:channel_size}\vspace{-2mm}
\end{table}

\begin{table}[t]
\centering
\resizebox{0.40\textwidth}{!}{
\begin{tabular}{lc|cc}
        \toprule
        \textbf{Type} & \textbf{Share} & OA (\%) & mAcc (\%) \\ %
        \midrule
        RoPE & - & 86.95 & 85.09 \\
        Learnable Embedding & - & 87.01 & 85.56 \\
        \midrule
        Linear & \small{\Checkmark} & 87.32 & 85.89 \\
        MLP & \small{\Checkmark} & 87.26 & 85.82 \\
        Linear & \XSolidBrush & 87.10 & 85.78 \\
        MLP & \XSolidBrush & 87.12 & 85.79 \\
        \bottomrule
        \end{tabular}
}
\caption{\small{\textbf{Ablation on positional embedding.} "Share" refers to learning a mapping function for all Mamba layers with the same channel size.}}\label{tab:pos_embed}\vspace{-6mm}
\end{table}

\subsection{Ablation Analysis and Visualization}
\label{sec:ablation}

\noindent
\textbf{Effect of serialization strategies.} The key to applying Mamba for point cloud modeling is transforming point clouds into point sequences. As shown in Table~\ref{tab:serialization}, we conduct the ablation experiment with different serialization strategies. Similar performance is achieved when all Mamba layers' inputs are serialized using a single order, whether it's z-order, Hilbert-order, or our proposed xyz-order. However, significant performance gains are observed when more serialization strategies are employed. When the three variants of our proposed consistent traverse serialization, namely "xyz", "yzx", and "zxy", are used together, PCM demonstrates a performance improvement of 0.17 OA and 0.11 mAcc compared to using only the "xyz" variant. When all six variants of consistent traverse serialization are combined, PCM shows a performance improvement of 0.39 OA and 0.51 mAcc. When all six variants of consistent traverse serialization, as well as "Hilbert," "z," and "z-trans" serialization strategies, are combined, PCM achieves a performance improvement of 0.49 OA and 0.54 mAcc. Different serialization strategies allow different Mamba layers to observe point clouds from different perspectives, resulting in more robust modeling of point cloud features.

\noindent
\textbf{Impact of channel size.} When the SSM-based method processes a token in the sequence, it relies solely on the hidden states and the input token, so the hidden states must have sufficient channel size to store global information. To investigate this, we conducted ablation studies on channel size, and the results are shown in Table~\ref{tab:channel_size}. When the channel size is reduced from 768 to 384, PCM exhibits a performance decay of 1.12 OA and 1.41 mAcc. A significant performance decay of 2.56 OA and 3.76 mAcc is observed when the channel size is reduced to 96. We then attempted to reduce the channel size of early stages, and the results show that excessively reducing the channel size of early stages (from 768 to 96) still leads to a performance decay of 1.36 OA and 2.2 mAcc. However, moderately reducing the channel size of the first two stages (from 768 to 384) only results in minor performance decay but can save significant computation. Therefore, in our final configuration, the first two stages adopt a channel size of 384, while the last two stages use a channel size of 768.\looseness=-1

\noindent
\textbf{Ablation on positional embedding.} We evaluate the impact of different positional encoding strategies, and the results are shown in Table~\ref{tab:pos_embed}. Initially, we experimented with rotary position embedding; however, it yielded the poorest performance. 
This is attributed to RoPE encoding solely the sequence order, unsuited for sparse and irregular point cloud data. 
Learnable positional embedding, a common practice in image sequence modeling, similarly encodes the sequence position and performs comparably to rotary positional encoding. Achieving favorable outcomes can be as straightforward as mapping point cloud spatial coordinates using a Linear layer as the positional encoding, resulting in a performance improvement of 0.37 OA and 0.7 mAcc compared to RoPE. 
Replacing the linear layer with a stronger MLP did not enhance performance. 
Nevertheless, the performance deteriorated due to overfitting when employing a separate Linear layer for each mamba layer.

\begin{table}[t]
\setlength{\tabcolsep}{1.0pt}
\begin{minipage}[t]{0.2\textwidth}
\centering
        \setlength{\tabcolsep}{0.9pt}
        \resizebox{!}{1.3cm}{
        \begin{tabular}{c|cc}
        \toprule
        \textbf{Prompts} & OA (\%) & mAcc (\%) \\ %
        \midrule
        0 & 86.64 & 84.70\\
        1 & 87.02 & 85.43\\
        3 & 86.88 & 84.78 \\
        6 & 87.47 & 86.17 \\
        12 & 87.13 & 85.28\\
        \bottomrule
        \end{tabular}
        }
        \caption{\small{\textbf{Ablation on numbers of order prompts.}}}\label{tab:order_prompts}
\end{minipage}\hfill\vspace{-2mm}
\begin{minipage}[t]{0.2\textwidth}
\centering
        \setlength{\tabcolsep}{1.5pt}
        \resizebox{!}{1.4cm}{
        \begin{tabular}{cc|cc}
        \toprule
        \textbf{K} & \textbf{Stride} & OA (\%) & mAcc (\%) \\ %
        \midrule
        0 & - & 79.32 & 76.25 \\
        4 & 1 & 84.66 & 82.72 \\
        8 & 1 & 86.95 & 85.06 \\
        12 & 1 & 87.37 & 85.96 \\
        24 & 1 & 87.09 & 85.10 \\
        24 & 2 & 85.39 & 83.55 \\
        24 & 4 & 83.73 & 81.61 \\
        \bottomrule
        \end{tabular}
        }
        \caption{\small{\textbf{Ablation on neighborhood points.}}}\label{tab:local_feat}
\end{minipage}\hfill\vspace{-3mm}
\end{table}

\noindent
\textbf{Order prompts.} To enhance the understanding of point cloud sequences by Mamba layers, we propose order prompts and conduct ablation experiments to validate their effectiveness, as shown in Table~\ref{tab:order_prompts}. When using only one order prompt, PCM demonstrates a performance improvement of 0.38 OA and 0.73 mAcc compared to not using any order prompt. Performance peaks when using six order prompts, resulting in a performance gain of 0.83 OA and 1.47 mAcc compared to no order prompt. However, further increasing the number of order prompts does not yield higher performance gains, although it still significantly outperforms not using any order prompt.\looseness=-1

\noindent
\textbf{Local features.} 3D point cloud data are sparse and have low semantic density, making local features crucial for understanding point clouds. We conducted ablation experiments on computing local features using different numbers of neighboring points, and the results are shown in Table~\ref{tab:local_feat}. When the number of neighboring points is set to 0, meaning no local features are computed, and only relying on Mamba layers to model the global features of the point cloud, PCM achieves an OA of 79.32 and a mAcc of 76.25. Using only four neighboring points to compute local features, PCM improves performance with an OA of 84.66 and a mAcc of 82.72, showing an increase of 5.34 OA and 6.47 mAcc compared to when local features are not used. PCM achieves the highest performance when using 12 points for computing local features. However, with a further increase in the number of neighboring points, performance decreases, indicating that the current local feature extraction mechanism, such as the geometric affine module in PointMLP, is not proficient at modeling the global features of point clouds. Therefore, combining local feature extraction modules and Mamba layers to model point clouds' local and global features is a promising approach.\looseness=-1

\vspace{-2mm}\section{Conclusion}
\label{sec:conclusion}

This paper introduces a Mamba-based point cloud network named Point Cloud Mamba, which, for the first time, outperforms the SOTA point-based method PointNeXt and transformer-based method PTv3. 
Point Cloud Mamba incorporates several novel techniques to help Mamba better model point cloud data. 
Firstly, we propose Consistent Traverse Serialization to convert 3D point cloud data into 1D point sequences that Mamba can handle, ensuring that neighboring points in the sequence are also spatially adjacent. 
Secondly, we aid Mamba in handling point sequences serialized in different orders by introducing order prompts containing sequence arrangement rules. 
Finally, we propose a simple yet effective positional encoding based on spatial coordinate mapping. 
Our Point Cloud Mamba achieves SOTA performance on the ScanObjectNN, ModelNet40, ShapeNetPart, and S3DIS datasets.
Adding a stronger local feature extractor, our method also outperforms previous STOA methods by a large margin on the S3DIS dataset.


%

\bibliography{AAAI/main}

\begin{thebibliography}{83}
\providecommand{\natexlab}[1]{#1}

\bibitem[{Armeni et~al.(2016)Armeni, Sener, Zamir, Jiang, Brilakis, Fischer, and Savarese}]{armeni20163d}
Armeni, I.; Sener, O.; Zamir, A.~R.; Jiang, H.; Brilakis, I.; Fischer, M.; and Savarese, S. 2016.
\newblock 3d semantic parsing of large-scale indoor spaces.
\newblock In \emph{CVPR}, 1534--1543.

\bibitem[{Behrouz and Hashemi(2024)}]{behrouz2024GMN}
Behrouz, A.; and Hashemi, F. 2024.
\newblock {Graph Mamba: Towards Learning on Graphs with State Space Models}.

\bibitem[{Boulch(2020)}]{boulch2020convpoint}
Boulch, A. 2020.
\newblock ConvPoint: Continuous convolutions for point cloud processing.
\newblock \emph{Computers \& Graphics}, 88: 24--34.

\bibitem[{Carion et~al.(2020)Carion, Massa, Synnaeve, Usunier, Kirillov, and Zagoruyko}]{detr}
Carion, N.; Massa, F.; Synnaeve, G.; Usunier, N.; Kirillov, A.; and Zagoruyko, S. 2020.
\newblock End-to-end object detection with transformers.
\newblock In \emph{ECCV}.

\bibitem[{Chang et~al.(2015)Chang, Funkhouser, Guibas, Hanrahan, Huang, Li, Savarese, Savva, Song, Su et~al.}]{chang2015shapenet}
Chang, A.~X.; Funkhouser, T.; Guibas, L.; Hanrahan, P.; Huang, Q.; Li, Z.; Savarese, S.; Savva, M.; Song, S.; Su, H.; et~al. 2015.
\newblock Shapenet: An information-rich 3d model repository.
\newblock \emph{arXiv:1512.03012}.

\bibitem[{Chen et~al.(2023{\natexlab{a}})Chen, Xia, Zang, Wang, and Li}]{chen2023decoupled}
Chen, B.; Xia, Y.; Zang, Y.; Wang, C.; and Li, J. 2023{\natexlab{a}}.
\newblock Decoupled local aggregation for point cloud learning.
\newblock \emph{arXiv preprint arXiv:2308.16532}.

\bibitem[{Chen et~al.(2023{\natexlab{b}})Chen, Wang, Yang, Yu, Yuan, and Yue}]{chen2023pointgpt}
Chen, G.; Wang, M.; Yang, Y.; Yu, K.; Yuan, L.; and Yue, Y. 2023{\natexlab{b}}.
\newblock PointGPT: Auto-regressively Generative Pre-training from Point Clouds.
\newblock \emph{arXiv:2305.11487}.

\bibitem[{Choy, Gwak, and Savarese(2019)}]{choy20194d}
Choy, C.; Gwak, J.; and Savarese, S. 2019.
\newblock 4d spatio-temporal convnets: Minkowski convolutional neural networks.
\newblock In \emph{CVPR}, 3075--3084.

\bibitem[{Dai et~al.(2017)Dai, Chang, Savva, Halber, Funkhouser, and Nie{\ss}ner}]{dai2017scannet}
Dai, A.; Chang, A.~X.; Savva, M.; Halber, M.; Funkhouser, T.; and Nie{\ss}ner, M. 2017.
\newblock Scannet: Richly-annotated 3d reconstructions of indoor scenes.
\newblock In \emph{CVPR}, 5828--5839.

\bibitem[{Dai and Nie{\ss}ner(2018)}]{dai20183dmv}
Dai, A.; and Nie{\ss}ner, M. 2018.
\newblock 3dmv: Joint 3d-multi-view prediction for 3d semantic scene segmentation.
\newblock In \emph{ECCV}, 452--468.

\bibitem[{Dosovitskiy et~al.(2021)Dosovitskiy, Beyer, Kolesnikov, Weissenborn, Zhai, Unterthiner, Dehghani, Minderer, Heigold, Gelly et~al.}]{dosovitskiy2020imageVIT}
Dosovitskiy, A.; Beyer, L.; Kolesnikov, A.; Weissenborn, D.; Zhai, X.; Unterthiner, T.; Dehghani, M.; Minderer, M.; Heigold, G.; Gelly, S.; et~al. 2021.
\newblock An image is worth 16x16 words: Transformers for image recognition at scale.
\newblock In \emph{ICLR}.

\bibitem[{Fang et~al.(2023)Fang, Li, Li, Buhmann, Loy, and Liu}]{fang2023explore}
Fang, Z.; Li, X.; Li, X.; Buhmann, J.~M.; Loy, C.~C.; and Liu, M. 2023.
\newblock Explore In-Context Learning for 3D Point Cloud Understanding.
\newblock \emph{NeurIPS}.

\bibitem[{Graham, Engelcke, and Van Der~Maaten(2018)}]{graham20183d}
Graham, B.; Engelcke, M.; and Van Der~Maaten, L. 2018.
\newblock 3d semantic segmentation with submanifold sparse convolutional networks.
\newblock In \emph{CVPR}, 9224--9232.

\bibitem[{Gu and Dao(2023)}]{gu2023mamba}
Gu, A.; and Dao, T. 2023.
\newblock Mamba: Linear-time sequence modeling with selective state spaces.
\newblock \emph{arXiv preprint arXiv:2312.00752}.

\bibitem[{Gu, Goel, and R\'e(2022)}]{gu2022efficiently}
Gu, A.; Goel, K.; and R\'e, C. 2022.
\newblock Efficiently Modeling Long Sequences with Structured State Spaces.
\newblock In \emph{ICLR}.

\bibitem[{Guo et~al.(2021)Guo, Cai, Liu, Mu, Martin, and Hu}]{guo2021pct}
Guo, M.-H.; Cai, J.-X.; Liu, Z.-N.; Mu, T.-J.; Martin, R.~R.; and Hu, S.-M. 2021.
\newblock Pct: Point cloud transformer.
\newblock In \emph{CVM}.

\bibitem[{He et~al.(2024)He, Cao, Yan, Li, Xie, Zhang, and Zhou}]{he2024pan}
He, X.; Cao, K.; Yan, K.; Li, R.; Xie, C.; Zhang, J.; and Zhou, M. 2024.
\newblock Pan-Mamba: Effective pan-sharpening with State Space Model.
\newblock \emph{arXiv preprint arXiv:2402.12192}.

\bibitem[{Hilbert and Hilbert(1935)}]{hilbert1935stetige}
Hilbert, D.; and Hilbert, D. 1935.
\newblock {\"U}ber die stetige Abbildung einer Linie auf ein Fl{\"a}chenst{\"u}ck.
\newblock \emph{Dritter Band: Analysis{\textperiodcentered} Grundlagen der Mathematik{\textperiodcentered} Physik Verschiedenes: Nebst Einer Lebensgeschichte}, 1--2.

\bibitem[{Hou et~al.(2022)Hou, Zhu, Ma, Loy, and Li}]{hou2022point}
Hou, Y.; Zhu, X.; Ma, Y.; Loy, C.~C.; and Li, Y. 2022.
\newblock Point-to-voxel knowledge distillation for lidar semantic segmentation.
\newblock In \emph{CVPR}, 8479--8488.

\bibitem[{Jiang et~al.(2022)Jiang, Lu, Zhao, Dazeley, and Wang}]{jiang2022mae3d}
Jiang, J.; Lu, X.; Zhao, L.; Dazeley, R.; and Wang, M. 2022.
\newblock Masked autoencoders in 3D point cloud representation learning.
\newblock \emph{arXiv:2207.01545}.

\bibitem[{Jiang et~al.(2023)Jiang, Yang, Shi, Golyanik, Dai, and Schiele}]{jiang2023self}
Jiang, L.; Yang, Z.; Shi, S.; Golyanik, V.; Dai, D.; and Schiele, B. 2023.
\newblock Self-supervised Pre-training with Masked Shape Prediction for 3D Scene Understanding.
\newblock In \emph{CVPR}, 1168--1178.

\bibitem[{Komarichev, Zhong, and Hua(2019)}]{komarichev2019acnn}
Komarichev, A.; Zhong, Z.; and Hua, J. 2019.
\newblock A-cnn: Annularly convolutional neural networks on point clouds.
\newblock In \emph{CVPR}.

\bibitem[{Lahoud et~al.(2022)Lahoud, Cao, Khan, Cholakkal, Anwer, Khan, and Yang}]{lahoud20223d}
Lahoud, J.; Cao, J.; Khan, F.~S.; Cholakkal, H.; Anwer, R.~M.; Khan, S.; and Yang, M.-H. 2022.
\newblock 3D Vision with Transformers: A Survey.
\newblock arXiv:2208.04309.

\bibitem[{Lai et~al.(2022)Lai, Liu, Jiang, Wang, Zhao, Liu, Qi, and Jia}]{lai2022stratified}
Lai, X.; Liu, J.; Jiang, L.; Wang, L.; Zhao, H.; Liu, S.; Qi, X.; and Jia, J. 2022.
\newblock Stratified Transformer for 3D Point Cloud Segmentation.
\newblock In \emph{CVPR}.

\bibitem[{Lang et~al.(2019)Lang, Vora, Caesar, Zhou, Yang, and Beijbom}]{lang2019pointpillars}
Lang, A.~H.; Vora, S.; Caesar, H.; Zhou, L.; Yang, J.; and Beijbom, O. 2019.
\newblock Pointpillars: Fast encoders for object detection from point clouds.
\newblock In \emph{CVPR}, 12697--12705.

\bibitem[{Li, Singh, and Grover(2024)}]{li2024mamba}
Li, S.; Singh, H.; and Grover, A. 2024.
\newblock Mamba-ND: Selective State Space Modeling for Multi-Dimensional Data.
\newblock \emph{arXiv preprint arXiv:2402.05892}.

\bibitem[{Li et~al.(2023)Li, Ding, Zhang, Yuan, Cheng, Jiangmiao, Chen, Liu, and Loy}]{li2023transformer}
Li, X.; Ding, H.; Zhang, W.; Yuan, H.; Cheng, G.; Jiangmiao, P.; Chen, K.; Liu, Z.; and Loy, C.~C. 2023.
\newblock Transformer-Based Visual Segmentation: A Survey.
\newblock \emph{arXiv pre-print}.

\bibitem[{Li et~al.(2018)Li, Bu, Sun, Wu, Di, and Chen}]{li2018pointcnn}
Li, Y.; Bu, R.; Sun, M.; Wu, W.; Di, X.; and Chen, B. 2018.
\newblock Pointcnn: Convolution on x-transformed points.
\newblock In \emph{NeurIPS}.

\bibitem[{Liang et~al.(2024)Liang, Zhou, Wang, Zhu, Xu, Zou, Ye, and Bai}]{liang2024pointmamba}
Liang, D.; Zhou, X.; Wang, X.; Zhu, X.; Xu, W.; Zou, Z.; Ye, X.; and Bai, X. 2024.
\newblock PointMamba: A Simple State Space Model for Point Cloud Analysis.
\newblock \emph{arXiv preprint arXiv:2402.10739}.

\bibitem[{Liu, Cai, and Lee(2022)}]{liu2022masked}
Liu, H.; Cai, M.; and Lee, Y.~J. 2022.
\newblock Masked discrimination for self-supervised learning on point clouds.
\newblock In \emph{ECCV}.

\bibitem[{Liu et~al.(2019)Liu, Fan, Xiang, and Pan}]{liu2019rscnn}
Liu, Y.; Fan, B.; Xiang, S.; and Pan, C. 2019.
\newblock Relation-shape convolutional neural network for point cloud analysis.
\newblock In \emph{CVPR}.

\bibitem[{Liu et~al.(2024)Liu, Tian, Zhao, Yu, Xie, Wang, Ye, and Liu}]{liu2024vmamba}
Liu, Y.; Tian, Y.; Zhao, Y.; Yu, H.; Xie, L.; Wang, Y.; Ye, Q.; and Liu, Y. 2024.
\newblock Vmamba: Visual state space model.
\newblock \emph{arXiv preprint arXiv:2401.10166}.

\bibitem[{Liu et~al.(2023)Liu, Yang, Tang, Yang, and Han}]{liu2023flatformer}
Liu, Z.; Yang, X.; Tang, H.; Yang, S.; and Han, S. 2023.
\newblock FlatFormer: Flattened Window Attention for Efficient Point Cloud Transformer.
\newblock In \emph{CVPR}, 1200--1211.

\bibitem[{Loshchilov and Hutter(2017)}]{loshchilov2017decoupled}
Loshchilov, I.; and Hutter, F. 2017.
\newblock Decoupled weight decay regularization.
\newblock \emph{arXiv preprint arXiv:1711.05101}.

\bibitem[{Ma, Li, and Wang(2024)}]{U-Mamba}
Ma, J.; Li, F.; and Wang, B. 2024.
\newblock U-Mamba: Enhancing Long-range Dependency for Biomedical Image Segmentation.
\newblock \emph{arXiv preprint arXiv:2401.04722}.

\bibitem[{Ma et~al.(2022)Ma, Qin, You, Ran, and Fu}]{ma2022pointmlp}
Ma, X.; Qin, C.; You, H.; Ran, H.; and Fu, Y. 2022.
\newblock Rethinking network design and local geometry in point cloud: A simple residual MLP framework.
\newblock In \emph{ICLR}.

\bibitem[{Morton(1966)}]{morton1966computer}
Morton, G.~M. 1966.
\newblock A computer oriented geodetic data base and a new technique in file sequencing.

\bibitem[{Pang et~al.(2022)Pang, Wang, Tay, Liu, Tian, and Yuan}]{pang2022point-mae}
Pang, Y.; Wang, W.; Tay, F.~E.; Liu, W.; Tian, Y.; and Yuan, L. 2022.
\newblock Masked autoencoders for point cloud self-supervised learning.
\newblock In \emph{ECCV}.

\bibitem[{Park et~al.(2023)Park, Lee, Kim, Xiong, and Kim}]{park2023self}
Park, J.; Lee, S.; Kim, S.; Xiong, Y.; and Kim, H.~J. 2023.
\newblock Self-positioning point-based transformer for point cloud understanding.
\newblock In \emph{ICCV}, 21814--21823.

\bibitem[{Qi et~al.(2017{\natexlab{a}})Qi, Su, Mo, and Guibas}]{qi2017pointnet}
Qi, C.~R.; Su, H.; Mo, K.; and Guibas, L.~J. 2017{\natexlab{a}}.
\newblock Pointnet: Deep learning on point sets for 3d classification and segmentation.
\newblock In \emph{CVPR}.

\bibitem[{Qi et~al.(2017{\natexlab{b}})Qi, Yi, Su, and Guibas}]{qi2017pointnet++}
Qi, C.~R.; Yi, L.; Su, H.; and Guibas, L.~J. 2017{\natexlab{b}}.
\newblock Pointnet++: Deep hierarchical feature learning on point sets in a metric space.
\newblock In \emph{NeurIPS}.

\bibitem[{Qian et~al.(2021)Qian, Hammoud, Li, Thabet, and Ghanem}]{qian2021assanet}
Qian, G.; Hammoud, H.; Li, G.; Thabet, A.; and Ghanem, B. 2021.
\newblock ASSANet: An Anisotropical Separable Set Abstraction for Efficient Point Cloud Representation Learning.
\newblock In \emph{NeurIPS}.

\bibitem[{Qian et~al.(2022)Qian, Li, Peng, Mai, Hammoud, Elhoseiny, and Ghanem}]{qian2022pointnext}
Qian, G.; Li, Y.; Peng, H.; Mai, J.; Hammoud, H.; Elhoseiny, M.; and Ghanem, B. 2022.
\newblock Pointnext: Revisiting pointnet++ with improved training and scaling strategies.
\newblock \emph{NeurIPS}.

\bibitem[{Ran, Liu, and Wang(2022)}]{ran2022surface}
Ran, H.; Liu, J.; and Wang, C. 2022.
\newblock Surface representation for point clouds.
\newblock In \emph{CVPR}.

\bibitem[{Robert, Raguet, and Landrieu(2023)}]{robert2023efficient}
Robert, D.; Raguet, H.; and Landrieu, L. 2023.
\newblock Efficient 3D semantic segmentation with superpoint transformer.
\newblock In \emph{ICCV}, 17195--17204.

\bibitem[{Ruan and Xiang(2024)}]{ruan2024vm}
Ruan, J.; and Xiang, S. 2024.
\newblock Vm-unet: Vision mamba unet for medical image segmentation.
\newblock \emph{arXiv preprint arXiv:2402.02491}.

\bibitem[{Sanghi(2020)}]{sanghi2020info3d}
Sanghi, A. 2020.
\newblock Info3d: Representation learning on 3d objects using mutual information maximization and contrastive learning.
\newblock In \emph{ECCV}, 626--642. Springer.

\bibitem[{Sauder and Sievers(2019)}]{sauder2019self}
Sauder, J.; and Sievers, B. 2019.
\newblock Self-supervised deep learning on point clouds by reconstructing space.
\newblock \emph{NIPS}, 32.

\bibitem[{Schult et~al.(2023)Schult, Engelmann, Hermans, Litany, Tang, and Leibe}]{Schult23ICRA}
Schult, J.; Engelmann, F.; Hermans, A.; Litany, O.; Tang, S.; and Leibe, B. 2023.
\newblock {Mask3D: Mask Transformer for 3D Semantic Instance Segmentation}.

\bibitem[{Shen et~al.(2018)Shen, Feng, Yang, and Tian}]{shen2018kcnet}
Shen, Y.; Feng, C.; Yang, Y.; and Tian, D. 2018.
\newblock Mining point cloud local structures by kernel correlation and graph pooling.
\newblock In \emph{CVPR}.

\bibitem[{Shi and Rajkumar(2020)}]{shi2020point}
Shi, W.; and Rajkumar, R. 2020.
\newblock Point-gnn: Graph neural network for 3d object detection in a point cloud.
\newblock In \emph{CVPR}, 1711--1719.

\bibitem[{Su et~al.(2024)Su, Ahmed, Lu, Pan, Bo, and Liu}]{su2024roformer}
Su, J.; Ahmed, M.; Lu, Y.; Pan, S.; Bo, W.; and Liu, Y. 2024.
\newblock Roformer: Enhanced transformer with rotary position embedding.
\newblock \emph{Neurocomputing}, 568: 127063.

\bibitem[{Sun et~al.(2023)Sun, Qing, Tan, and Xu}]{sun2022superpoint}
Sun, J.; Qing, C.; Tan, J.; and Xu, X. 2023.
\newblock Superpoint Transformer for 3D Scene Instance Segmentation.
\newblock \emph{AAAI}.

\bibitem[{Sun et~al.(2024)Sun, Rao, Lu, and Yan}]{sun2024x}
Sun, S.; Rao, Y.; Lu, J.; and Yan, H. 2024.
\newblock X-3D: Explicit 3D Structure Modeling for Point Cloud Recognition.
\newblock In \emph{CVPR}, 5074--5083.

\bibitem[{Thomas et~al.(2019)Thomas, Qi, Deschaud, Marcotegui, Goulette, and Guibas}]{thomas2019kpconv}
Thomas, H.; Qi, C.~R.; Deschaud, J.-E.; Marcotegui, B.; Goulette, F.; and Guibas, L.~J. 2019.
\newblock Kpconv: Flexible and deformable convolution for point clouds.
\newblock In \emph{ICCV}.

\bibitem[{Thomas et~al.(2024)Thomas, Tsai, Barfoot, and Zhang}]{thomas2024kpconvx}
Thomas, H.; Tsai, Y.-H.~H.; Barfoot, T.~D.; and Zhang, J. 2024.
\newblock KPConvX: Modernizing Kernel Point Convolution with Kernel Attention.
\newblock In \emph{CVPR}, 5525--5535.

\bibitem[{Uy et~al.(2019)Uy, Pham, Hua, Nguyen, and Yeung}]{uy2019scanobjectnn}
Uy, M.~A.; Pham, Q.-H.; Hua, B.-S.; Nguyen, T.; and Yeung, S.-K. 2019.
\newblock Revisiting point cloud classification: A new benchmark dataset and classification model on real-world data.
\newblock In \emph{ICCV}.

\bibitem[{Vaswani et~al.(2017)Vaswani, Shazeer, Parmar, Uszkoreit, Jones, Gomez, Kaiser, and Polosukhin}]{vaswani2017attention}
Vaswani, A.; Shazeer, N.; Parmar, N.; Uszkoreit, J.; Jones, L.; Gomez, A.~N.; Kaiser, {\L}.; and Polosukhin, I. 2017.
\newblock Attention is all you need.
\newblock In \emph{NeurIPS}.

\bibitem[{Wang et~al.(2019)Wang, Huang, Hou, Zhang, and Shan}]{wang2019graph}
Wang, L.; Huang, Y.; Hou, Y.; Zhang, S.; and Shan, J. 2019.
\newblock Graph attention convolution for point cloud semantic segmentation.
\newblock In \emph{CVPR}, 10296--10305.

\bibitem[{Wang(2023)}]{wang2023octformer}
Wang, P.-S. 2023.
\newblock OctFormer: Octree-based Transformers for 3D Point Clouds.
\newblock \emph{SIGGRAPH}.

\bibitem[{Wang et~al.(2024)Wang, Fang, Li, Li, Chen, and Liu}]{wang2023skeleton}
Wang, X.; Fang, Z.; Li, X.; Li, X.; Chen, C.; and Liu, M. 2024.
\newblock Skeleton-in-Context: Unified Skeleton Sequence Modeling with In-Context Learning.
\newblock \emph{CVPR}.

\bibitem[{Wang and Solomon(2019)}]{wang2019deep}
Wang, Y.; and Solomon, J.~M. 2019.
\newblock Deep closest point: Learning representations for point cloud registration.
\newblock In \emph{ICCV}, 3523--3532.

\bibitem[{Wu, Qi, and Fuxin(2019)}]{wu2019pointconv}
Wu, W.; Qi, Z.; and Fuxin, L. 2019.
\newblock Pointconv: Deep convolutional networks on 3d point clouds.
\newblock In \emph{CVPR}.

\bibitem[{Wu et~al.(2024)Wu, Jiang, Wang, Liu, Liu, Qiao, Ouyang, He, and Zhao}]{wu2023ptv3}
Wu, X.; Jiang, L.; Wang, P.-S.; Liu, Z.; Liu, X.; Qiao, Y.; Ouyang, W.; He, T.; and Zhao, H. 2024.
\newblock Point Transformer V3: Simpler, Faster, Stronger.
\newblock In \emph{CVPR}.

\bibitem[{Wu et~al.(2022)Wu, Lao, Jiang, Liu, and Zhao}]{wu2022ptv2}
Wu, X.; Lao, Y.; Jiang, L.; Liu, X.; and Zhao, H. 2022.
\newblock Point transformer V2: Grouped Vector Attention and Partition-based Pooling.
\newblock In \emph{NeurIPS}.

\bibitem[{Wu et~al.(2023{\natexlab{a}})Wu, Tian, Wen, Peng, Liu, Yu, and Zhao}]{wu2023towards}
Wu, X.; Tian, Z.; Wen, X.; Peng, B.; Liu, X.; Yu, K.; and Zhao, H. 2023{\natexlab{a}}.
\newblock Towards large-scale 3d representation learning with multi-dataset point prompt training.
\newblock \emph{arXiv preprint arXiv:2308.09718}.

\bibitem[{Wu et~al.(2023{\natexlab{b}})Wu, Wen, Liu, and Zhao}]{wu2023masked}
Wu, X.; Wen, X.; Liu, X.; and Zhao, H. 2023{\natexlab{b}}.
\newblock Masked scene contrast: A scalable framework for unsupervised 3d representation learning.
\newblock In \emph{CVPR}, 9415--9424.

\bibitem[{Wu et~al.(2015)Wu, Song, Khosla, Yu, Zhang, Tang, and Xiao}]{wu2015ModelNet40}
Wu, Z.; Song, S.; Khosla, A.; Yu, F.; Zhang, L.; Tang, X.; and Xiao, J. 2015.
\newblock 3d shapenets: A deep representation for volumetric shapes.
\newblock In \emph{CVPR}.

\bibitem[{Xiang et~al.(2021)Xiang, Zhang, Song, Yu, and Cai}]{xiang2021curvenet}
Xiang, T.; Zhang, C.; Song, Y.; Yu, J.; and Cai, W. 2021.
\newblock Walk in the cloud: Learning curves for point clouds shape analysis.
\newblock In \emph{ICCV}.

\bibitem[{Xie et~al.(2020)Xie, Gu, Guo, Qi, Guibas, and Litany}]{xie2020pointcontrast}
Xie, S.; Gu, J.; Guo, D.; Qi, C.~R.; Guibas, L.; and Litany, O. 2020.
\newblock Pointcontrast: Unsupervised pre-training for 3d point cloud understanding.
\newblock In \emph{ECCV}.

\bibitem[{Xing et~al.(2024)Xing, Ye, Yang, Liu, and Zhu}]{xing2024segmamba}
Xing, Z.; Ye, T.; Yang, Y.; Liu, G.; and Zhu, L. 2024.
\newblock Segmamba: Long-range sequential modeling mamba for 3d medical image segmentation.
\newblock \emph{arXiv preprint arXiv:2401.13560}.

\bibitem[{Xu et~al.(2021{\natexlab{a}})Xu, Ding, Zhao, and Qi}]{xu2021paconv}
Xu, M.; Ding, R.; Zhao, H.; and Qi, X. 2021{\natexlab{a}}.
\newblock Paconv: Position adaptive convolution with dynamic kernel assembling on point clouds.
\newblock In \emph{CVPR}.

\bibitem[{Xu et~al.(2021{\natexlab{b}})Xu, Zhang, Zhou, Xu, Qi, and Qiao}]{xu2021GDANet}
Xu, M.; Zhang, J.; Zhou, Z.; Xu, M.; Qi, X.; and Qiao, Y. 2021{\natexlab{b}}.
\newblock Learning geometry-disentangled representation for complementary understanding of 3d object point cloud.
\newblock In \emph{AAAI}.

\bibitem[{Yan et~al.(2022)Yan, Gao, Zheng, Zheng, Zhang, Cui, and Li}]{yan20222dpass}
Yan, X.; Gao, J.; Zheng, C.; Zheng, C.; Zhang, R.; Cui, S.; and Li, Z. 2022.
\newblock 2dpass: 2d priors assisted semantic segmentation on lidar point clouds.
\newblock In \emph{ECCV}. Springer.

\bibitem[{Yang, Xing, and Zhu(2024)}]{yang2024vivim}
Yang, Y.; Xing, Z.; and Zhu, L. 2024.
\newblock Vivim: a video vision mamba for medical video object segmentation.
\newblock \emph{arXiv preprint arXiv:2401.14168}.

\bibitem[{Yang et~al.(2023)Yang, Guo, Xiong, Liu, Pan, Wang, Tong, and Guo}]{yang2023swin3d}
Yang, Y.-Q.; Guo, Y.-X.; Xiong, J.-Y.; Liu, Y.; Pan, H.; Wang, P.-S.; Tong, X.; and Guo, B. 2023.
\newblock Swin3D: A Pretrained Transformer Backbone for 3D Indoor Scene Understanding.
\newblock \emph{arXiv preprint arXiv:2304.06906}.

\bibitem[{Yi et~al.(2016)Yi, Kim, Ceylan, Shen, Yan, Su, Lu, Huang, Sheffer, and Guibas}]{yi2016shapenetpart}
Yi, L.; Kim, V.~G.; Ceylan, D.; Shen, I.-C.; Yan, M.; Su, H.; Lu, C.; Huang, Q.; Sheffer, A.; and Guibas, L. 2016.
\newblock A scalable active framework for region annotation in 3d shape collections.
\newblock In \emph{TOG}.

\bibitem[{Yu et~al.(2022)Yu, Tang, Rao, Huang, Zhou, and Lu}]{yu2021pointbert}
Yu, X.; Tang, L.; Rao, Y.; Huang, T.; Zhou, J.; and Lu, J. 2022.
\newblock Point-BERT: Pre-Training 3D Point Cloud Transformers with Masked Point Modeling.
\newblock In \emph{CVPR}.

\bibitem[{Zhang et~al.(2021)Zhang, Girdhar, Joulin, and Misra}]{zhang2021DepthContras}
Zhang, Z.; Girdhar, R.; Joulin, A.; and Misra, I. 2021.
\newblock Self-supervised pretraining of 3d features on any point-cloud.
\newblock In \emph{ICCV}.

\bibitem[{Zhao et~al.(2021)Zhao, Jiang, Jia, Torr, and Koltun}]{point_transformer}
Zhao, H.; Jiang, L.; Jia, J.; Torr, P.~H.; and Koltun, V. 2021.
\newblock Point Transformer.
\newblock In \emph{ICCV}.

\bibitem[{Zhu et~al.(2023)Zhu, Yang, Wu, Huang, Zhang, He, He, Zhao, Shen, Qiao et~al.}]{zhu2023ponderv2}
Zhu, H.; Yang, H.; Wu, X.; Huang, D.; Zhang, S.; He, X.; He, T.; Zhao, H.; Shen, C.; Qiao, Y.; et~al. 2023.
\newblock Ponderv2: Pave the way for 3d foundataion model with a universal pre-training paradigm.
\newblock \emph{arXiv preprint arXiv:2310.08586}.

\bibitem[{Zhu et~al.(2024)Zhu, Liao, Zhang, Wang, Liu, and Wang}]{zhu2024vision}
Zhu, L.; Liao, B.; Zhang, Q.; Wang, X.; Liu, W.; and Wang, X. 2024.
\newblock Vision mamba: Efficient visual representation learning with bidirectional state space model.
\newblock \emph{arXiv preprint arXiv:2401.09417}.

\bibitem[{Zhu et~al.(2021)Zhu, Zhou, Wang, Hong, Ma, Li, Li, and Lin}]{zhu2021cylindrical}
Zhu, X.; Zhou, H.; Wang, T.; Hong, F.; Ma, Y.; Li, W.; Li, H.; and Lin, D. 2021.
\newblock Cylindrical and asymmetrical 3d convolution networks for lidar segmentation.
\newblock In \emph{CVPR}, 9939--9948.

\end{thebibliography}


\newpage 
\clearpage


\newpage

\begin{table}[t]
\centering
\resizebox{0.47\textwidth}{!}{
\begin{tabular}{l|ccc}
\toprule
\textbf{Method}  & OA & mAcc & mIOU\\ %
\midrule
PointNeXt-XL~\cite{qian2022pointnext} & - & - & 71.5 \\
Strat. Trans.~\cite{lai2022stratified} & - & - & 74.3 \\
PTv1~\cite{point_transformer} & - & - & 70.6 \\
PTv2~\cite{wu2022ptv2} & - & - & 75.4 \\
DeLA~\cite{chen2023decoupled} & 91.6 & 82.8 & 74.7\\
\midrule
PCM-Tiny\dag~(ours) & \textbf{91.8} & \textbf{84.2} & \textbf{75.5}\\
\bottomrule
\end{tabular}
}
\caption{\textbf{3D semantic segmentation in ScanNet.} \dag~indicates using DeLA~\cite{chen2023decoupled} blocks as the additional local feature extractor.}
\label{tab:scannet}
\end{table}

\begin{table}[t]
\centering
\resizebox{0.47\textwidth}{!}{
\begin{tabular}{l|ccc|cc}
\toprule
\multirow{2}{*}{\textbf{Method}}  & Params.  & FLOPs & Throughput  & \multicolumn{2}{c}{ScanObjectNN} \\  
 & M & G & ins./sec. & OA & mAcc\\
\midrule
PointMLP~\cite{ma2022pointmlp}& 13.2 & 31.4 & \textbf{447} & 85.4 & 83.9 \\
PCM-Tiny & \textbf{6.9} & \textbf{11.0} & 256 & 86.9 & 85.0 \\
PCM & 34.2 & 45.0 & 148 & \textbf{88.1} & \textbf{86.6} \\

\bottomrule
\end{tabular}
}
\caption{\textbf{Comparison of parameters, computational complexity, and inference speed.}}
\label{tab:flops}
\end{table}

\begin{table}[t]
\centering
        \resizebox{0.47\textwidth}{!}{
        \begin{tabular}{l|cc}
        \toprule
        \textbf{Points} & OA (\%) & mAcc (\%) \\ %
        \midrule
        \{1024-1024-1024-1024\} & 87.35 & 85.71 \\
        \{1024-512-256-128\} & 87.20 & 85.54 \\
        \{512-256-128-128\} & 86.95 & 85.32 \\
        \{512-256-128-64\} & 86.68 & 85.12 \\
        \bottomrule
        \end{tabular}
        }
\caption{\textbf{Ablation on points downsampling.} The number of points at different stages is listed within \{\} and connected with -.}\label{tab:downsampling}
\end{table}

\begin{table*}[t]
\centering
\caption{Comparison with naive mamba-based architecture on ScanObjectNN.}
\label{tab:scanobjectnn_supp}
\resizebox{1.0\textwidth}{!}{
\begin{tabular}{l|cc|ccccccccccccccc}
\toprule
\textbf{Method}  & OA & mAcc & bag & bin & box & cabinet & chair & desk & display & door & shelf & table & bed & pillow & sink & sofa & toilet \\ %
\midrule
PointMamba~\cite{liang2024pointmamba} & 82.5 & - & - & - & - & - & - & - & - & - & - & - & - & - & - & - & - \\
Naive arch. (ours) & 85.2 & 83.0 & 66.3 & 85.4 & 61.7 & 87.1 & 94.4 & 80.0 & 87.3 & 94.3 & 88.0 & 75.6 & 84.6 & 85.7 & 77.5 & 92.9 & 84.7\\
PCM (ours) & \textbf{88.0} & \textbf{85.9} & 65.1 & 90.5 & 79.0 & 87.9 & 98.0 & 82.0 & 89.7 & 95.7 & 91.7 & 75.9 & 80.0 & 86.7 & 83.3 & 96.7 & 85.9 \\
\bottomrule
\end{tabular}
}
\end{table*}

\begin{table*}[t]
\centering
\caption{Comparison with naive mamba-based architecture on ModelNet40.}
\label{tab:modelnet40_supp}
\resizebox{1.0\textwidth}{!}{
\begin{tabular}{l|cccccccccccccccc} 
\midrule
\multirow{6}{*}{Naive arch.} & OA & mAcc & airplane & bathtub & bed & bench & bookshelf & bottle & bowl & car & chair & cone & cup  & curtain & desk & door\\
 & 91.9 & 88.3 & 100.0 & 96.0 & 100.0 & 70.0 & 98.0 & 94.0 & 90.0 & 98.0 & 98.0 & 90.0 & 60.0 & 95.0 & 89.5 & 90.0 \\
 & dresser & flower\_pot &glass\_box &guitar &keyboard &lamp  &laptop &mantel &monitor &night\_stand&person  &piano  &plant &radio  & range\_hood & sink\\
 & 83.7 & 15.0 & 94.0 & 100.0 & 100.0 & 85.0 &  100.0 & 97.0 & 100.0 & 82.6 & 95.0 & 95.0 & 83.0 & 70.0 & 93.0 & 90.0 \\
 & sofa & stairs &stool  &table &tent &toilet &tv\_stand &vase &wardrobe &xbox \\
 & 100.0 & 85.0 & 85.0 & 88.0 & 95.0 & 100.0 & 82.0 & 86.0 & 80.0 & 80.0 \\
\hline
 \multirow{6}{*}{PCM (ours)} & OA & mAcc & airplane & bathtub & bed & bench & bookshelf & bottle & bowl & car & chair & cone & cup  & curtain & desk & door\\
 & 93.4 & 90.7 & 100.0 & 96.0 & 99.0 & 75.0 & 100.0 & 98.0 & 95.0 & 99.0 & 98.0 & 100.0 & 80.0 & 95.0 & 89.5 & 90.0 \\
 & dresser & flower\_pot &glass\_box &guitar &keyboard &lamp  &laptop &mantel &monitor &night\_stand&person  &piano  &plant &radio  & range\_hood & sink\\
 & 86.1 & 10.0 & 95.0 & 100.0 & 100.0 & 95.0 & 100.0 & 96.0 & 99.0 & 82.6 & 90.0 & 91.0 & 90.0 & 90.0 & 98.8 & 95.0 \\
 & sofa & stairs &stool  &table &tent &toilet &tv\_stand &vase &wardrobe &xbox \\
 &  100.0 & 95.0 & 80.0 & 92.0 & 95.0 & 99.0 & 88.0 & 83.0 & 75.0 & 90.0 \\
\bottomrule
\end{tabular}
}
\end{table*}

\begin{table*}[t]
\centering
\caption{Comparison with naive mamba-based architecture on ShapeNetPart.}
\label{tab:shapenetpart_supp}
\resizebox{1.0\textwidth}{!}{
\begin{tabular}{l|cc|cccccccccccccccc}
\toprule
\textbf{Method}  & Ins. nIoU & Cls. mIoU & airplane & bag & cap & car & chair & earphone & guitar & knife & lamp & laptop & motorbike & mug & pistol & rocket & skateboard & table \\  
\midrule
PointMamba~\cite{liang2024pointmamba} & 85.8 & 83.9 & - & -& -& -& -& -& -& -& -& -& -& -& -& -& -& - \\
Naive arch. (ours) & 86.6 & 84.8 & 84.9 & 88.4 & 86.0 & 81.9 & 91.8 & 79.0 & 92.3 & 87.9 & 85.5 & 95.8 & 76.6 & 95.9 & 83.8 & 66.6 & 77.1 & 83.4 \\
PCM (ours) & \textbf{87.0} & \textbf{85.3} & 86.2 & 87.2 & 89.3 & 85.2 & 92.1 & 81.4 & 92.4 & 88.3 & 85.0 & 96.5 & 79.2 & 96.0 & 86.0 & 62.8 & 77.0 & 83.3 \\
\bottomrule
\end{tabular}
}
\end{table*}

\begin{table*}[t]
\centering
\caption{Comparison with the single serialization strategy on ShapeNetPart.}
\label{tab:serialization_supp}
\resizebox{1.0\textwidth}{!}{
\begin{tabular}{l|cc|cccccccccccccccc}
\toprule
\textbf{Method}  & Ins. nIoU & Cls. mIoU & airplane & bag & cap & car & chair & earphone & guitar & knife & lamp & laptop & motorbike & mug & pistol & rocket & skateboard & table \\  
\midrule
Single & 86.6 & 84.6 & 85.2 & 86.0 & 88.4 & 81.2 & 92.1 & 80.9 & 91.9 & 87.9 & 85.1 & 95.9 & 78.3 & 96.1 & 84.7 & 59.8 & 77.8 & 83.0 \\
Multiple & \textbf{87.0} & \textbf{85.3} & 86.2 & 87.2 & 89.3 & 85.2 & 92.1 & 81.4 & 92.4 & 88.3 & 85.0 & 96.5 & 79.2 & 96.0 & 86.0 & 62.8 & 77.0 & 83.3 \\
\bottomrule
\end{tabular}
}
\end{table*}

In this supplementary, we present the implementation details, more experiment and visualization results, and further works.

\section{Implementation Details}

\noindent\textbf{Additional local feature extractor.} Local features are very important for the semantic segmentation of point clouds. To further enhance PCM's capability for local feature modeling, we introduce additional DeLA~\cite{chen2023decoupled} blocks into PCM. Specifically, we cascade 4 DeLA blocks with PCM. The point cloud first passes through DeLA blocks to obtain point features, and then these point features are used as input to PCM for local and global modeling. The additional DeLA blocks and PCM are trained from scratch on semantic segmentation datasets without pre-training. 

\noindent\textbf{Experiment Setup.} We train PCM using the AdamW optimizer~\cite{loshchilov2017decoupled} with an initial learning rate of 1e-4, employing a Cosine Decay and a weight decay of 1e-4. For ScanObjectNN, ModelNet40 and ShapeNetPart datasets, we perform warmup for 5 epochs and use a batch size of 32. For the S3DIS dataset, we perform warmup for 5\% iterations and use a batch size of 16. We train PCM for 250 epochs on the ScanObjectNN and ModelNet40 datasets, for 300 epochs on ShapeNetPart, and 3000 epochs on S3DIS. For ScanObjectNN and ModelNet40, we followed the PointNeXt using 1024 points, randomly sampled during training, and using farthest point sampling during testing. For ShapeNetPart, 2,048 randomly sampled points with normals were used as input for training and testing. For S3DIS, 30,000 randomly sampled points were used as input for training. Following PointNeXt~\cite{qian2022pointnext}, PCM employs multi-step learning rate decay during training on ShapeNetPart, decaying at epochs 210 and 270, with a decay rate of 0.5. The experimental settings for PCM-Tiny are identical to PCM on all datasets. All ablation experiments are conducted using PCM as the default architecture, implemented on the ScanObjectNN dataset with training shortened to 125 epochs. Apart from this adjustment, all other settings are identical to the main experiment.

\section{More Experiment Results.}
\label{sec:more_exp}

\noindent\textbf{3D semantic segmentation in ScanNet dataset.} As shown in Figure~\ref{tab:scannet}, PCM achieved 75.5 mIoU on the ScanNet~\cite{dai2017scannet} benchmark, surpassing DeLA by 0.8 mIoU, 1.4 mAcc, and 0.2 OA. Compared to PointNeXt-XL, our proposed PCM significantly outperformed it by 4.0 mIoU.

\noindent
\textbf{Comparison with naive mamba-based architecture.} 
We compare our proposed PCM with the naive Mamba-based architecture on ScanObjectNN~\cite{uy2019scanobjectnn}, ModelNet40~\cite{wu2015ModelNet40}, and ShapeNetPart~\cite{yi2016shapenetpart} to validate the effectiveness of our design. 
The comparison results on ScanObjectNN are shown in Table~\ref{tab:scanobjectnn_supp}. 
It is worth noting that even our proposed naive Mamba-based architecture outperformed the contemporaneous work PointMamba~\cite{liang2024pointmamba} 2.7 OA due to the combination of local modeling and global modeling. 
However, thanks to our proposed consistent traverse serialization strategy, order prompt, simple positional embedding, and more reasonable architecture settings, our PCM still surpasses the naive architecture with 2.8 OA and 2.9 mAcc on ScanObjectNN. 
Moreover, the accuracy in almost all categories is higher than in the naive Mamba-based architecture. 
The comparison results on ModelNet40 are shown in Tab.~\ref{tab:modelnet40_supp}. PCM also outperforms the naive Mamba-based architecture with 1.5 OA and 2.4 mAcc.

The comparison of part segmentation performance on the ShapeNetPart dataset is shown in Tab.~\ref{tab:shapenetpart_supp}. 
The naive Mamba-based architecture surpasses the contemporaneous work PointMamba with 0.8 Ins. mIoU and 0.9 Cls. mIoU. 
PCM also exceeds the naive architecture with 0.4 Ins. mIoU and 0.5 Cls. mIoU.

\noindent
\textbf{Point downsampling.} Point cloud data exhibit significant redundancy; thus, appropriate point downsampling can substantially reduce computational costs with minimal loss in performance. We experimented with several downsampling schemes, and the results are shown in Table~\ref{tab:downsampling}. Downsampling by a factor of 2 for all stages except the first stage resulted in only a slight performance decrease of 0.15 OA and 0.17 mAcc while significantly reducing computation. However, excessive downsampling, leaving only 64 points for the last stage, led to a performance drop of 0.67 OA and 0.59 mAcc. We ultimately adopted a downsampling strategy of line 2 for PCM.

\noindent
\textbf{Comparison of the multiple serialization strategy with the single serialization strategy.} In the main paper, we compared the impact of the multiple serialization strategy and single serialization strategy on point cloud classification performance. 
The multiple serialization strategy yielded improvements of 0.42 OA and 0.87 mAcc compared to the single serialization strategy. 
As shown in Table~\ref{tab:serialization_supp}, on ShapeNetPart, the multiple serialization strategy also led to performance enhancements of 0.4 Ins. mIoU and 0.7 Cls. mIoU in part segmentation.

\noindent
\textbf{Comparison of the Parameters, FLOPs, and Throughput.} We summarize our proposed PCM's parameter, computational complexity, and throughput, as shown in Table~\ref{tab:flops}. 
Our proposed PCM-Tiny outperformed PointMLP 1.5 OA and 1.1 mAcc with only 52\% parameters and 35\% computational complexity. 
PCM surpassed PointMLP~\cite{ma2022pointmlp} 2.7 OA and 2.7 mAcc with a larger amount of parameters and computational complexity. 
However, due to multiple reorderings, the throughput of PCM is not advantageous; even though PCM-Tiny has fewer parameters than PointMLP, its throughput is still lower than PointMLP. 

\begin{figure}[t]
\centering
\includegraphics[width=0.5\textwidth]{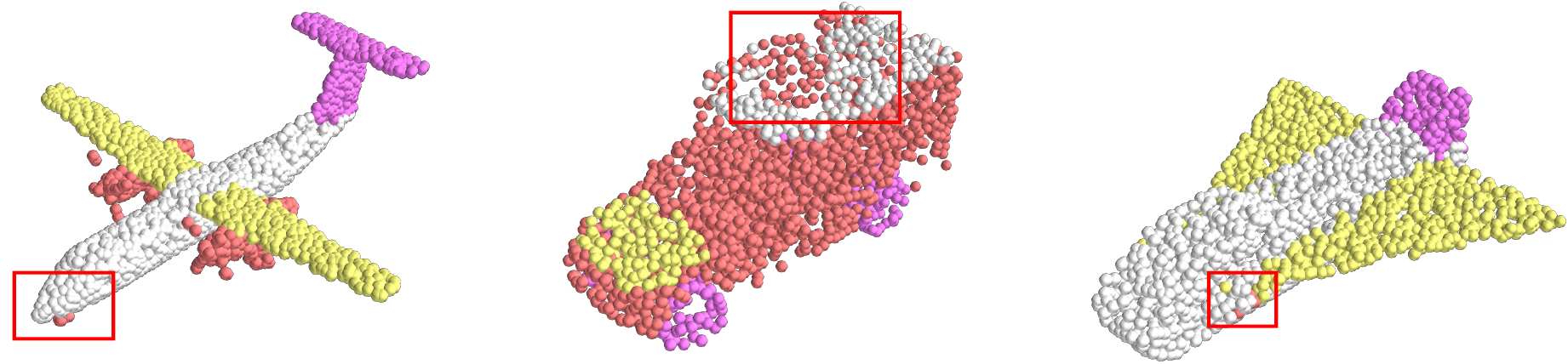}
\caption{\textbf{The failure cases of PCM.} Incorrect areas are highlighted by red rectangles.}
\label{fig:failure}
\end{figure}

\begin{figure}[t]
\centering
\includegraphics[width=0.48\textwidth]{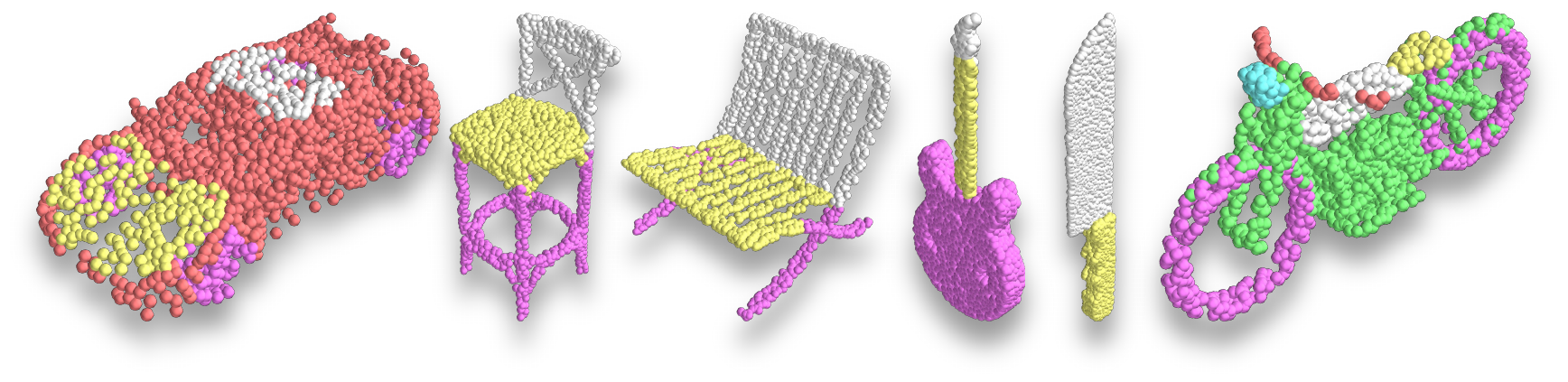}
\caption{\textbf{The visualization results of part segmentation on the ShapeNetPart dataset.}}
\label{fig:visualization}
\vspace{-1em}
\end{figure}

\begin{figure}[t]
\centering
\includegraphics[width=0.5\textwidth]{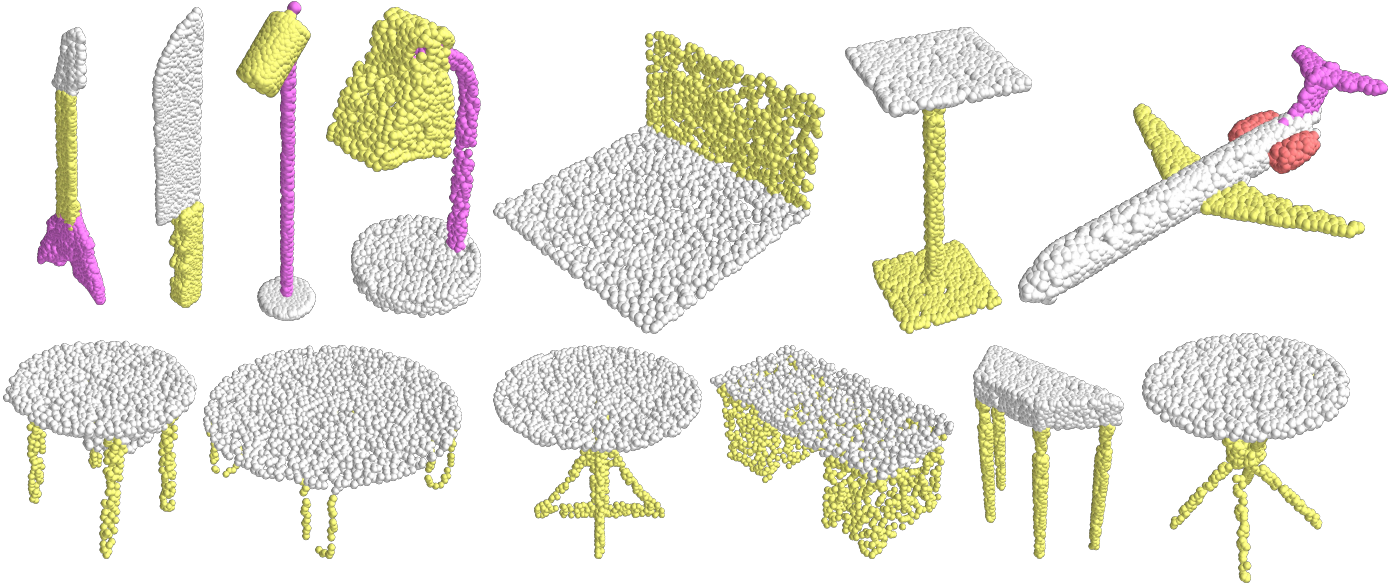}
\caption{\textbf{More visualization results.}}
\label{fig:cases}
\end{figure}

\section{Visualization Results}
\label{sec:more_visual}

\noindent
\textbf{Failure cases.} Figure~\ref{fig:failure} illustrates some instances of PCM failure. 
For example, on the left side, PCM performs poorly for certain smaller object parts, often misclassifying them as parts of larger, similar ones. 
PCM is also susceptible to issues when the point cloud has numerous missing points, as evidenced by the car in the middle.

\noindent
\textbf{More visualization results.} In Figures~\ref{fig:cases} and \ref{fig:visualization}, we present more visualization results. 
Even when dealing with elongated or flattened objects, PCM still achieves good results. 
This demonstrates that serializing the point cloud into a point sequence and then using Mamba to model global features is feasible and effective, even when the point cloud is distributed irregularly in space.

\section{Further works.}
\label{sec:further_works}

In future work, we will focus on how to utilize Mamba for the global modeling of large-scale point cloud scenes. Since Mamba employs scan-based computation during training to enhance parallelism, which incurs quadratic computational complexity, it is not feasible to directly process the whole point cloud with Mamba during training. However, during testing, the entire point cloud is often processed at once, creating a substantial gap between training and testing and thereby limiting the performance of Mamba-based methods. We will explore strategies to bridge this gap, such as through scalable serialization methods.

\noindent
\textbf{Limitations and future work directions.}
PCM successfully introduces Mamba into point cloud analysis and surpasses modern point-based methods like PointNeXt. However, there are still some limitations that need to be addressed. For large-scale point clouds (i.e., $\geq$ 100k points), such as in the S3DIS dataset, Mamba struggles to handle such long sequences during training due to the scan-based computational approach used to accelerate training. 
Therefore, it is necessary to crop the input point clouds, but this introduces a gap between training and inference for global modeling architecture PCM. 
In addition, how to better combine local feature extractors with PCM is also worth trying.
Moreover, there are still several directions to explore when adopting PCM in out-door point cloud scene, where the point inputs are huge and more complex.
We will focus on addressing these challenges in our future work.

\end{document}